\documentclass[fleqn,10pt]{wlscirep}
\usepackage[utf8]{inputenc}
\usepackage[T1]{fontenc}

\usepackage{multirow}  % For better tables
\usepackage{xcolor}  % For colored text
\usepackage{booktabs} % Better tables
\usepackage{adjustbox} % Somethign something

\newcommand{\cgreen}[2]{
    \textcolor[rgb]{0.,#1,0.}{#2}
}
\newcommand{\cred}[2]{
    \textcolor[rgb]{#1,0.,0.}{#2}
}

\usepackage{subcaption}  % For subfigures

\usepackage{todonotes}

\title{Enhancing the Reliability of Medical AI through Expert-guided Uncertainty Modeling}

\author[1,2*+]{Aleksei Khalin}
\author[1,2+]{Ekaterina Zaychenkova}
\author[2,3]{Aleksandr Yugay}
\author[3]{Andrey Goncharov}
\author[1,2]{Sergey Korchagin}
\author[3,4]{Alexey Zaytsev}
\author[2,5]{Egor Ershov}
\affil[1]{Kharkevich Institute for Information Transmission Problems of Russian Academy of Sciences, Moscow, Russia}
\affil[2]{Moscow Insitute of Physics and Technology, Dolgoprudny, Russia}
\affil[3]{Skolkovo Institute of Science and Technology, Moscow, Russia}
\affil[4]{Sber, Moscow, Russia}
\affil[5]{Artificial Intelligence Research Institute, Moscow, Russia}

\affil[*]{Correspondence: khalin.av@color.iitp.ru}

\affil[+]{These authors contributed equally to this work}


\begin{abstract}
Artificial intelligence (AI) systems accelerate medical workflows and improve diagnostic accuracy in healthcare, serving as second-opinion systems.
However, the unpredictability of AI errors poses a significant challenge, particularly in healthcare contexts, where mistakes can have severe consequences. 
A widely adopted safeguard is to pair predictions with uncertainty estimation, enabling human experts to focus on high-risk cases while streamlining routine verification. 
Current uncertainty estimation methods, however, remain limited, particularly in quantifying aleatoric uncertainty, which arises from data ambiguity and noise.
To address this, we propose a novel approach that leverages disagreement in expert responses to generate targets for training machine learning models.
These targets are used in conjunction with standard data labels to estimate two components of uncertainty separately, as given by the law of total variance, via a two-ensemble approach, as well as its lightweight variant.
We validate our method on binary image classification, binary and multi-class image segmentation, and multiple-choice question answering. 
Our experiments demonstrate that incorporating expert knowledge can enhance uncertainty estimation quality by $9\%$ to $50\%$ depending on the task, making this source of information invaluable for the construction of risk-aware AI systems in healthcare applications.
\end{abstract}
\begin{document}

\flushbottom
\maketitle
\thispagestyle{empty}

\section*{Introduction}

Computer-based systems, especially those that leverage artificial intelligence (AI), offer significant improvements in speed and quality in healthcare applications. 
AI can be used to accelerate analysis, provide diagnostic recommendations, mitigate human-factor errors, highlight critical details and complex cases, and identify pathologies that may fall outside the primary diagnostic focus~\cite{rajpurkar2022ai}. 
By automating routine tasks, such as entering results into laboratory information systems, digital automation enables the reallocation of human expertise to tasks where it remains irreplaceable.
This is especially crucial in large laboratories, where the volume of routine tests can lead to fatigue-related errors, and most reactions are sufficiently simple for automated analysis. 

The present problem is AI's susceptibility to unpredictable errors, whereas reliability is paramount in medicine.  
Because errors in medical processes can result in severe health damage, including fatal outcomes, many countries legally mandate that only a human make the final decision, taking responsibility for the consequences~\cite{briganti2020artificial}.  
A standard approach to tackle this problem is to provide an AI-based opinion to a human expert.
However, excessive revision is inefficient.
Thus, a key challenge is to keep necessary oversight without over-reliance on AI.  

Uncertainty estimation (UE) is a common way to address this issue when the system's prediction is accompanied by its uncertainty in the output.
This addition reflects the complexity of the task: the greater the uncertainty, the more likely the error, and thus the greater the need for human intervention.
This framework enables a medical expert to focus on high-risk, complex cases while allowing rapid verification of simpler ones.

Recent studies advocate that an effective approach to quantifying uncertainty is to split it into two types depending on their origin and estimate them separately due to their fundamentally different nature~\cite{gawlikowski2023survey}.  
% Aleatoric, or data, uncertainty (AU) is caused by ambiguity in the data that happens because of stochasticity and errors during data generation, inherent noise.
% AU is an irreducible value.
Epistemic, or model, uncertainty (EU) arises from the lack of knowledge about the system being described, and can be reduced with more data or better models.
Aleatoric, or data, uncertainty (AU) arises from inherent variability in data generation stemming from randomness, sensor noise, and imprecision in data-collection methods; unlike epistemic uncertainty, it is irreducible even with additional data.
% Epistemic, or model, uncertainty (EU) arises from the scarcity of a training set resulting in the lack of model knowledge.
% EU can be reduced by improving of the training set. 
% This is why AU estimation is more significant according to ref: https://arxiv.org/pdf/1703.04977

\begin{figure}[t]
    \centering
    \includegraphics[height=4.2in]{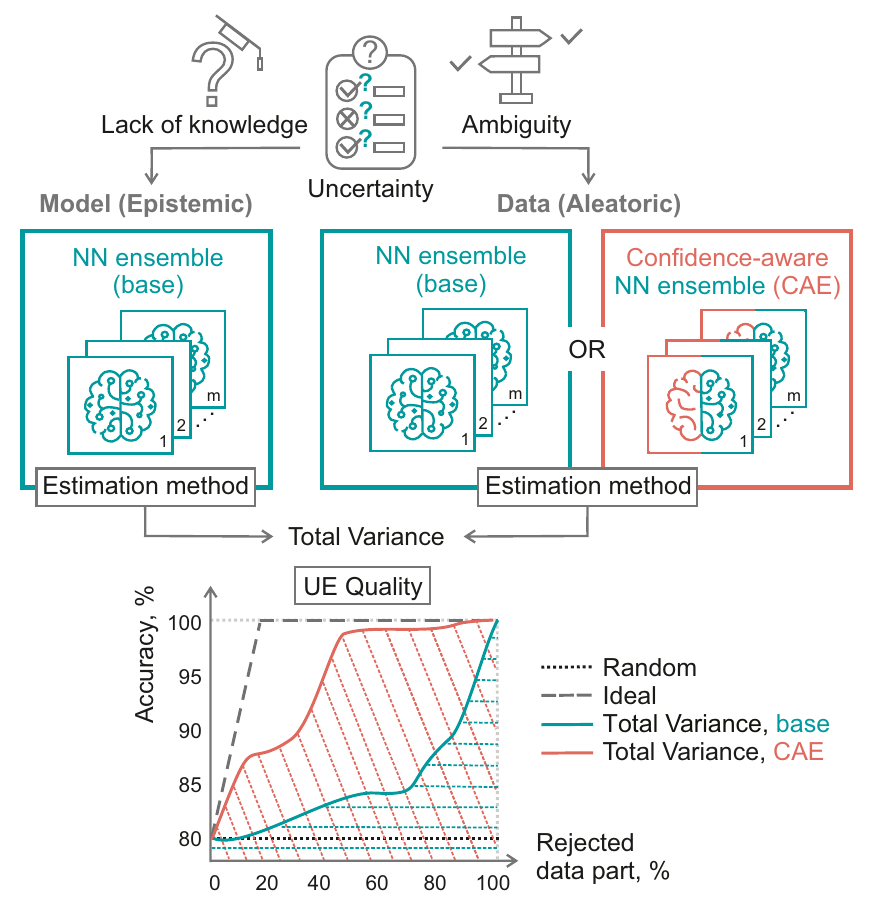}
    \caption{Integration of expert knowledge into aleatoric uncertainty estimation provides total UE quality enhancement.}
    \label{fig:frontal}
\end{figure}

% However, existing UE methods remain imperfect.
Existing UE methods remain imperfect, with aleatoric uncertainty estimation being particularly challenging due to the absence of a reliable ground truth and inherent over-confidence of modern neural networks~\cite{guo2017calibration}.
% Aleatoric component estimation is especially challenging due to the absence of a reliable ground truth and inherent over-confidence of modern neural networks~\cite{guo2017calibration}.
Motivated by this observation, we propose a novel framework that integrates expert knowledge into the AU estimation.
To accurately decompose the total uncertainty, we use the law of total variance~\cite{loeve2017probability}.  
We estimate EU using a neural network ensemble trained on categorical (``hard'') labels.
This ensemble is then finetuned on experts' votes (``soft'' labels) to estimate AU, thereby incorporating knowledge about data complexity while preserving the models' classification performance.
The core concept of our approach is illustrated in Fig.~\ref{fig:frontal}.

Furthermore, to enhance practical utility, we introduce a simplified one-ensemble alternative method. 
We demonstrate that it achieves a comparable performance ($34\%$ improvement over the baseline approach, versus $55\%$ for the two-ensemble method) while offering greater training and inference efficiency.

% Our theoretical foundation of the proposed approach focuses on classification (here we treat segmentation as pixel-wise classification) tasks since most medical problems, such as diagnostics, pathology identification in MRI, CT, and histopathological images, and interpretation of reaction results, can be effectively framed as classification problems.

\paragraph{Our main contributions are as follows:}
\begin{enumerate}
    \item We provide evidence across diverse digital medicine problems that expert confidence evaluations improve the quality of uncertainty estimates. 
    % Expert markup upgrades binary labels to cool soft labels
    Predictions from experts are used to generate ``soft'' labels that capture aleatoric uncertainty and can be used for either direct uncertainty estimation or training of neural networks.
    % We investigate the applicability of responses from human experts and large language models as digital experts.
    % In both cases, the uncertainty related to data quality and complexity (aleatoric) becomes more adequate. 
    % \item To incorporate expert annotations as an additional source of knowledge, we propose a novel approach to aleatoric uncertainty estimation for machine learning models.
    The law of total variance allows for the splitting of the uncertainty into two components, leading to the estimation of both aleatoric and epistemic components separately using different types of labels.
    This approach is efficient, easy to implement, and theoretically justified.
    \item 
    We introduce a combination of two ensembles: the ensemble of neural networks to predict ``hard'' (binary) labels and a confidence-aware ensemble trained on soft labels, that estimates epistemic and aleatoric uncertainty, respectively.
    % Their combination via law of total variance is our method.
    \item Extensive experiments show that this method outperforms state-of-the-art approaches to uncertainty estimation across four various medical tasks; namely, the proposed solution achieves a $9\%$ improvement in multiple-choice question answering on the PubMedQA dataset, a $50\%$ improvement in image classification on the BloodyWell dataset, a $7\%$ improvement in binary image segmentation on the LIDC-IDRI dataset and a $49\%$ improvement in multiclass image segmentation on the RIGA dataset, compared to the second best solution.
    % In binary image classification (LIDC-IDRI dataset), our approach shows comparable performance to other UE techniques. 
    \item Our work includes wide-ranging ablations and a sensitivity study. To mitigate ensemble cost, we analyse the dependence of uncertainty quality on ensemble size. We then assess use cases without expert markup and with limited training time.
    % \item To address the high computational cost of ensemble training, we study the dependence of uncertainty estimation quality on the number of ensemble models; experiments show that our solution requires fewer computational resources to achieve equal or better results than other approaches.
    % We also propose a one-ensemble approach that reduces training time while maintaining satisfactory quality. 
\end{enumerate}

\section*{Related work}

% TODO: add no less than three papers from NatCom

% \todo[inline]{TODO by Aleksandr. Maybe combine with intro?}

% \todo{introduce the structure of the section in the first paragraph and focus not on baselines (we need to present them and motivate the usage of them, but it is not the only thing that we need)}

% Machine learning systems that lack an estimate of prediction reliability can be insufficient for safe and reliable decision-making in healthcare~\cite{lindenmeyer2025towards}.
% Without such information, clinicians cannot distinguish between routine cases and those in which the model may fail, leading to high-impact errors. 
% Uncertainty estimation techniques address this gap by providing a measure of prediction confidence, enabling identification of cases that require expert verification, detection of distributional shift, and prevention of silent failures~\cite{wahid2024artificial}. 
Studies in musculoskeletal and neurological applications further highlight that appropriate UE strategies improve clinician trust by providing calibrated confidence levels and clearer communication of model limitations~\cite{vahdani2025towards,molchanova2025explainability}. 
Recent scoping reviews in radiotherapy and medical imaging indicate that, although UE is wastly recognized as essential for clinical deployment, many systems still rely on simple techniques and often do not provide a robust decomposition of different uncertainty sources~\cite{wahid2024artificial,huang2024review}.
Such decomposition can be helpful because it allows more accurate modeling of separate components and meets diverse needs across different applications~\cite{budd2021survey,ma2025survey}.

In machine learning, uncertainty is commonly categorized into aleatoric and epistemic~\cite{kendall2017uncertainties,hullermeier2021aleatoric}. 
% Aleatoric uncertainty captures the inherent variability in the data, such as overlapping class distributions, ambiguous visual patterns, or noisy measurements that cannot be reduced even with more data.
% Epistemic uncertainty reflects uncertainty about the model parameters; it arises from limited training data, distributional shift, or model misspecification, and can in principle be reduced with additional information~\cite{hullermeier2021aleatoric}.
A major practical challenge is that ground truth labels for these uncertainty types are typically unavailable~\cite{hullermeier2021aleatoric}, making direct evaluation impossible; therefore, most methods rely on indirect assessments, such as measuring predictive calibration, evaluating performance under distributional shifts, or using synthetic data where uncertainty components can be controlled.

Bayesian methods provide a principled framework for uncertainty estimation by treating model parameters as random variables~\cite{denker1990transforming}. 
In this view, epistemic uncertainty arises from the posterior distribution over model parameters, reflecting the model's limited knowledge in regions of the input space not well represented in the training data. 
Aleatoric uncertainty is captured by the conditional distribution of predictions given a specific set of parameters, reflecting the inherent noise or variability in the data. 
Exact Bayesian inference is analytically and computationally intractable in deep learning because it requires integrating over a high-dimensional parameter space. 
So, several approximation techniques are applied to make uncertainty estimation feasible in practice:
Variational inference approximates the posterior over parameters with a simpler distribution~\cite{hinton1993keeping}, Monte Carlo dropout estimates uncertainty primarily by capturing epistemic variability through repeated stochastic forward passes~\cite{gal2016dropout}, Laplace approximation models the posterior as a normal distribution around the optimum~\cite{immer2021improving}, and Markov Chain Monte Carlo (MCMC) generates ensembles of models by stochastically sampling parameter space~\cite{welling2011bayesian}. 

Although Bayesian approaches rigorously quantify uncertainty, practical applications often adopt simpler alternatives. 
For example, ensemble methods approximate Bayesian uncertainty by training multiple models and measuring prediction variability, capturing epistemic uncertainty akin to sampling from a posterior over model parameters.
Diversity within an ensemble of deep learning models can be introduced through random initialization~\cite{lakshminarayanan2017simple}.
Alternative approaches consider different neural network architectures for ensemble members~\cite{herron2020ensembles}, training on different data subsets~\cite{bishop2006pattern}, or training-time augmentations~\cite{nanni2019data}. 

To account for aleatoric uncertainty, test-time augmentation perturbs inputs at inference and measures the resulting variability in predictions~\cite{wang2018test}. 
Post-hoc calibration techniques, such as temperature scaling, Platt scaling, and isotonic regression, adjust predicted probabilities to better match empirical frequencies, thereby providing a more faithful estimate of aleatoric uncertainty without altering the underlying model parameters~\cite{guo2017calibration}. 
More advanced, adaptive calibration schemes further refine confidence estimates in a context-dependent and fine-grained manner, enhancing the reliability of probability predictions in practical applications~\cite{tomani2022parameterized,balanya2024adaptive}.
However, in the absence of ground truth, even for proper scoring rules, the amount of data required for reliable uncertainty estimation without confidence labels remains enormous.

To the best of our knowledge, no methods that leverage expert knowledge to improve uncertainty estimation exist.
However, diverse ways to integrate prior human knowledge into machine learning are shown to improve different capabilities of machine learning models~\cite{von2021informed}.
Human-in-the-loop approaches involve experts during model development, such as in Expert-Augmented Machine Learning, where expert-assigned risk assessments regularize the model~\cite{mosqueira2023human,gennatas2020expert}. 
A mature ecosystem of annotation platforms~\cite{douglas2023data} facilitates large-scale data collection by implementing protocols to optimize limited expert time and ensure high-quality output~\cite{albert2023comparing,hasan2024boosting}.
After obtaining such labels, they can be incorporated directly as input features, as in Knowledge-Infused Representations~\cite{biermann2022knowledge} or KINN~\cite{chattha2019kinn}.
Modern generative AI models also emerge as an alternative to human annotators~\cite{bazarova2025hallucination,pangakis2025keeping}.
% Soft labels, probabilistic annotations that help models better capture uncertainty, especially in small-sample or imbalanced settings~\cite{nguyen2014learning,de2025learning}, exist.
Thus, for uncertainty estimation, experts can also provide complementary knowledge to improve the reliability and interpretability of machine learning systems. 

Despite substantial progress, a gap remains, as no approaches jointly perform uncertainty estimation and leverage expert confidence annotations.
We argue for frameworks that fuse aleatoric-epistemic decomposition with expert-informed signals to deliver uncertainty estimates for the model predictions.
Working in this direction would benefit trustworthy decision support in high-stakes domains such as healthcare.

\section*{Methods: enriching uncertainty estimation with expert knowledge}

% \todo[inline]{Introduce what is uncertainty. Polish the text below}

% Neural networks are routinely used in medicine for prediction, with inputs ranging from images to natural language.
% To enforce risk-aware decision-making, we want a neural network to be equipped with an uncertainty estimate that correlates strongly with the probability of model error.
% However, vanilla uncertainty estimates, if available, appear unreliable.
% Specifically, they show limited usability for making abstention decisions based on uncertainty: if we score all examples with an uncertainty estimator and reject the given share of the most uncertain ones for manual inspection, the performance would be limited.
% Such a procedure helps to improve the reliability of AI-augmented systems, simultaneously minimizing the amount of manual work completed by domain experts. 

The key obstacles in uncertainty estimation arise both in defining it and training an uncertainty-aware model.
Firstly, the uncertainty originates from two sources: a data-related component (low data quality or high data complexity, aleatoric) and a model-related component (lack of knowledge about a specific model, epistemic). 
Secondly, a typical dataset includes only zero-one marks for each data example, lacking nuanced intermediate values.
We design a method that considers both sources of uncertainty, aiming at further improving a data uncertainty estimate using expert non-binary confidence labels.

% \todo[inline]{This section would benefit greatly from better analysis of existing uncertainty decomposition approaches.}

% Several approaches exist for decomposing the uncertainty of machine learning models into epistemic and aleatoric components.
There are two widely used frameworks for decomposing the uncertainty: the entropy-based approach~\cite{hullermeier2021aleatoric}, where predictive entropy is interpreted as total uncertainty of predictions, and is decomposed in terms of mutual information;
and the variance-based approach~\cite{depeweg2018decomposition,sale2024label}, where total uncertainty is quantified as total variance of model predictions, and epistemic and aleatoric components are derived from the law of total variance.
Although classification problems typically adopt the entropy-based approach, it has been shown not to always work as expected~\cite{wimmer2023quantifying}.
Recent work, on the other hand, has demonstrated potential benefits of applying the variance-based approach for classification~\cite{duan2024evidential}.
Moreover, it has been proved that total variance better satisfies the theoretical requirements naturally imposed on uncertainty measures~\cite{sale2024label}.
For these reasons, we adopt the variance-based approach for this paper and review it below.

\subsection*{Variance-based approach to uncertainty estimation}

The general idea behind the variance-based approach is that a target value $y$, which the machine learning model is trained to predict, is modeled as a random variable, and the total uncertainty of the predictions is defined as the variance of this random variable. 
The variance is then decomposed into two summands, and each of them is considered to relate to one of the uncertainty components.

Consider a binary classification problem: train a model $p_{\theta}(x)$ with parameters $\theta$ that predicts which of two classes, that we call class $0$ or class $1$, a sample $x$ should be assigned to.
Class label $y \in \{0, 1\}$ is modeled as a random variable with a Bernoulli distribution conditioned on $x$: $y \mid x \sim \mathrm{Bern}(p(x))$ with $p(x) = \mathbb{P}(y = 1 \mid x)$.
A model $p_\theta(x)$ is then trained to predict the probability of an object $x$ to have class $y = 1$: $p_\theta(x) = \mathbb{P}(y = 1 \mid x, \theta) \approx \mathbb{P}(y = 1 \mid x)$.

The variance-based approach quantifies total uncertainty as total variance of the class label $y \mid x$: $\mathrm{TU}(y) = \mathrm{Var}(y \mid x)$.
The law of total variance~\cite{loeve2017probability} allows us to view total uncertainty as a sum of two components:
\begin{equation}
    \mathrm{Var}(y\mid x) = \mathrm{Var}_{\theta \sim \mathbb{P}(\theta)}[\mathbb{E}(y\mid x, \theta)] + \mathbb{E}_{\theta\sim\mathbb{P}(\theta)}[\mathrm{Var}(y\mid x, \theta)].
    \label{eq:total-variance}
\end{equation}

The first term $\mathrm{Var}_{\theta \sim \mathbb{P}(\theta)}[\mathbb{E}(y\mid x, \theta)]$ relates to the epistemic uncertainty and quantifies how much predictions can vary between models with different parameter sets $\theta$ sampled from $\mathbb{P}(\theta)$.
Typically, this distribution becomes sharper with larger data size, as it corresponds to the posterior density $\mathbb{P}(\theta) = \mathbb{P}(\theta | D)$ for a training sample $D$.
Thus, as more data are introduced during training, predictions become more consistent across models, reducing variance.
% This term relates to the epistemic uncertainty (EU) of classification: $\mathrm{EU}(y) = \mathrm{Var}_{\theta \sim \mathbb{P}(\theta)}[\mathbb{E}(y\mid x, \theta)]$.

The second term $\mathbb{E}_{\theta\sim\mathbb{P}(\theta)}[\mathrm{Var}(y\mid x, \theta)]$ reflects randomness inherent to the classification process, which stems from noise and ambiguity in data.
It is irreducible, even with additional data, and therefore can be viewed as a measure of aleatoric uncertainty (AU): $\mathrm{AU}(y) = \mathbb{E}_{\theta\sim\mathbb{P}(\theta)}[\mathrm{Var}(y\mid x, \theta)]$.

Monte Carlo methods enable approximation of both components by sampling multiple parameter sets $\{\theta_i\}_{i = 1}^K$, i.e., training an ensemble of $K$ different models, yielding the following formulas used in practice:
\begin{align}
    \mathrm{Var}_{\theta\sim \mathbb{P}(\theta)} \left[\mathbb{E}(y \mid x, \theta)\right] &\approx \frac{1}{K}\sum_{i=1}^{K}\left(p_{\theta_i}(x) - \overline{p}_{\theta}(x)\right)^2, \label{eq:epistemic} \\ 
    \mathbb{E}_{\theta\sim \mathbb{P}(\theta)}\left[\mathrm{Var}(y \mid x, \theta)\right] &\approx \frac{1}{K}\sum_{i=1}^{K} p_{\theta_i}(x)\left(1 - p_{\theta_i}(x)\right), \label{eq:aleatoric}
\end{align}
where $\overline{p}_{\theta}(x) = \frac1K\sum_{i = 1}^K p_{\theta_i}(x)$ denotes the ensemble mean prediction.

This paradigm can be extended to the scenario of multi-class classification with class label $y \in \{1, \ldots, N\}$.
If we denote $y_n = [y = n]$ for each $n \in \{1, \ldots, N\}$, then combined uncertainty from all classes can be viewed as the sum of respective uncertainties for each class in the one-versus-all manner:
\begin{equation}
    \mathrm{TU}(y) = \sum_{n = 1}^N \mathrm{TU}(y_n), \hspace{3em}
    \mathrm{EU}(y) = \sum_{n = 1}^N \mathrm{EU}(y_n), \hspace{3em}
    \mathrm{AU}(y) = \sum_{n = 1}^N \mathrm{AU}(y_n).
\end{equation}
Such aggregation may appear counterintuitive, as classes in multi-class segmentation are not independent of each other, and the summation of per-class uncertainties is not guaranteed to be mathematically meaningful.
We provide a rationale for this approach in Supplementary Methods.

\subsection*{Inconsistency in expert annotations as a measure of aleatoric uncertainty}

When multiple experts annotate data, there is inevitably disagreement among their responses for some samples.
This inconsistency directly reflects the irreducible ambiguity in the labeling task -- precisely what aleatoric uncertainty captures.
Thus, an aggregation of expert responses could provide an approximation of $p(x)$ better fitted for evaluating the aleatoric component given by equation~\eqref{eq:aleatoric}.
It is possible to leverage expert knowledge to produce a ``soft'' label $\tilde{y}$ that captures information on aleatoric uncertainty while retaining discriminative performance.

The use of such an estimate in practice, however, is often infeasible: cost and speed of producing expert confidence labels for a new sample remain, and even increase, when multiple persons are involved.
As an alternative, we fine-tune a model to predict such labels instead of ``hard'' labels $y$.
It can then be used to better account for the presence of aleatoric uncertainty in its predictions.
This idea lies at the heart of our proposed approach, which we present below.

\subsection*{Training confidence-aware ensembles based on expert annotations}
\begin{figure}[ht]
    \centering
    \includegraphics[height=4.2in]{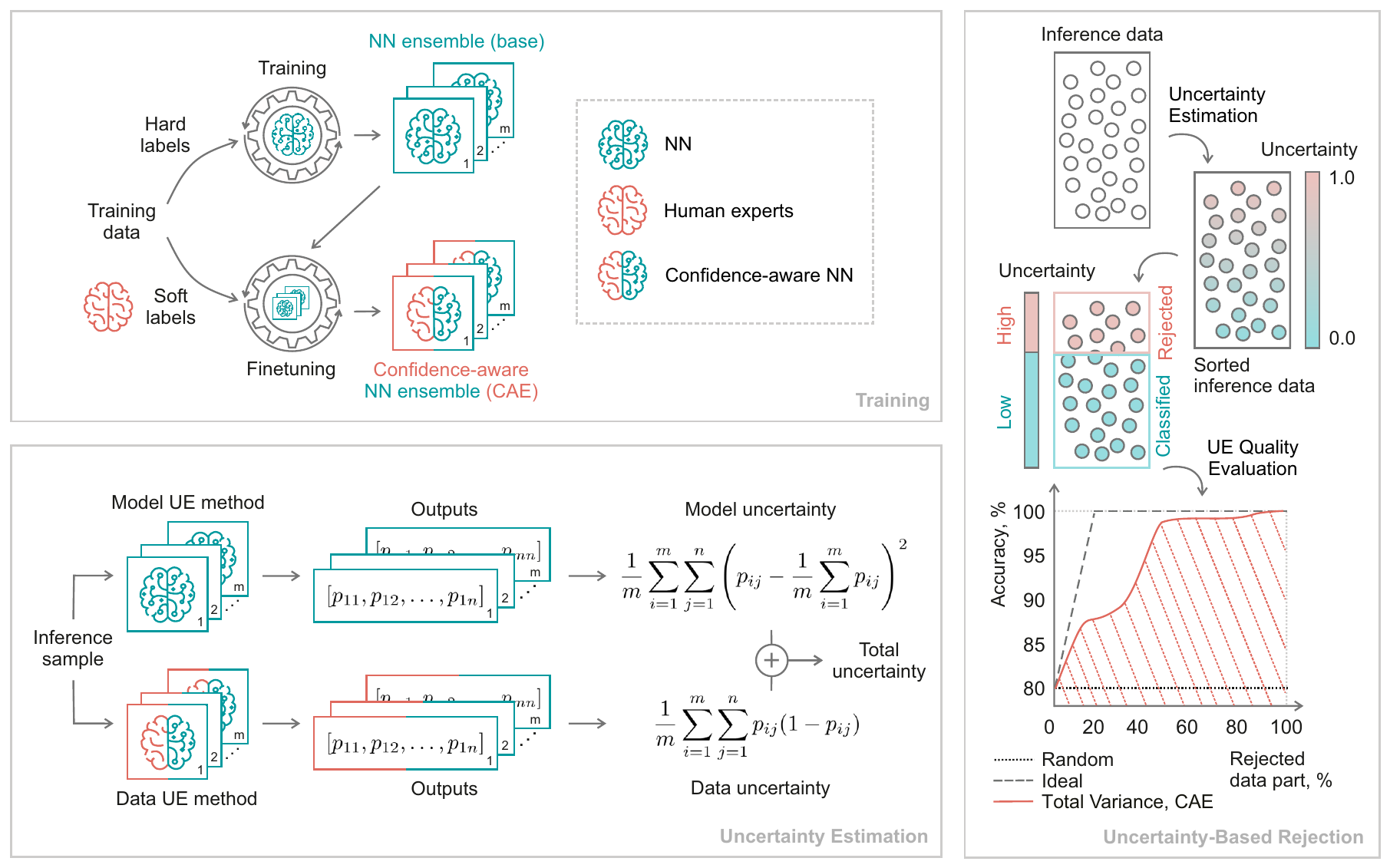}
    \caption{Overview of the proposed method. First, an ensemble of models is trained to predict ground truth labels from the dataset. Next, a different ensemble is created via fine-tuning on ``soft'' labels acquired from expert annotations. The two ensembles are then used to estimate epistemic and aleatoric uncertainty, respectively, given by equations~\eqref{eq:epistemic}~and~\eqref{eq:aleatoric}. Finally, we evaluate the performance of algorithms with different uncertainty thresholds and construct rejection curves.}
    \label{fig:methods}
\end{figure}

Our solution consists of three stages illustrated in Fig.~\ref{fig:methods}. 
We train two separate ensembles to predict epistemic and aleatoric uncertainty, and discuss the specifics below.

\subsubsection*{Ensemble training}

First, a base ensemble of $M$ models is trained to provide an estimate of epistemic uncertainty.
To do so, each model is trained to predict ground truth labels extracted from the dataset.
To provide variability in model responses, weights for each model instance are initialized randomly\cite{lakshminarayanan2017simple} and data augmentation is used during training\cite{bishop2006pattern} where possible.
To streamline the process, the architecture choice is the same across all ensemble models and depends only on the task.

Next, a different ensemble of $M$ models, that we refer to as a confidence-aware ensemble (CAE), is produced to estimate the aleatoric component of uncertainty.
Each model of a new ensemble is first initialized with the weights of a model from the base ensemble.
That is done to ensure the stability of training.
Then, the models are trained to predict ``soft'' labels generated from expert annotations.
In general, a simple averaging of responses is enough to produce good results, as we show in our experiments.
In certain tasks, however, it is beneficial to apply more complex techniques (see Methods for details).

\subsubsection*{Uncertainty estimation via a combination of ensembles}

To estimate the final uncertainty value per sample, we generate two sets of predictions $\{p_{ij}\}_{i,j = 1}^{M, N}$ and $\{\tilde p_{ij}\}_{i, j = 1}^{M, N}$ from a base ensemble and a CAE respectively.
Here, $p_{ij}$ (or $\tilde p_{ij}$) is the confidence of model $i$ ($\tilde{i}$) that a sample belongs to class $j$. 
%$p_{ij} \geq 0$ for all $i, j$ and, for any $i \in \{1, \ldots, M\}$, $\sum_{j = 1}^N p_{ij} = 1$.
The first set of predictions is used to evaluate epistemic uncertainty via~\eqref{eq:epistemic}, and the second is substituted in~\eqref{eq:aleatoric} to evaluate the aleatoric uncertainty.
We sum these two values to produce an estimation of the total variance for a given sample.

In principle, it is mathematically possible to use the CAE to evaluate epistemic uncertainty as well.
However, since epistemic uncertainty stems directly from the lack of model knowledge, it is best estimated by a base ensemble trained directly to solve the task at hand.
% We also partially confirm this claim in our experiments.

\subsubsection*{Rejection curves for comparison of uncertainty estimation methods}

% \todo[inline]{A brief mention of other existing metrics like calibration metrics would be great.}
Several methods exist for measuring the quality of uncertainty estimation.
A common approach in literature is to relate uncertainty estimation to calibration of machine learning models: a model is said to be \textit{well-calibrated} if it outputs lower uncertainty estimates for samples with lower error rates.
Quality of uncertainty is then quantified in terms of calibration errors, namely Expected Calibration Error~\cite{posocco2021estimating}, which bins predictions and measures the difference between uncertainty estimates and error rates in each bin;
Maximum Calibration Error~\cite{naeini2015obtaining}, which measures maximum deviation between uncertainty and error rate;
as well as other similar measures~\cite{nixon2019measuring,pmlr-v80-kumar18a}.
However, using calibration comes with various challenges.
Calibration curves are not easily extended to tasks beyond classification, as they rely on predictions either being correct or incorrect.
Additionally, the aforementioned estimators of calibration error often lack interpretability.

For these reasons, we adopt rejection curves~\cite{nadeem2009accuracy} (also known as risk coverage curves) to compare methods of uncertainty estimation as they reflect a real-life use case of uncertainty estimation, where a second-opinion system is allowed to abstain from classifying a fraction of the samples and request a double-check by an expert.
To construct a rejection curve, all samples in the test part of the dataset are ordered based on the estimated uncertainty value.
For a fixed \textit{rejection rate} $r \in [0, 1]$, a fraction $r$ of samples with the highest uncertainty values are discarded from the test set, and model performance is evaluated by a target metric on the remaining data, resulting in the value $\mathrm{M}(r)$.
The choice of metric is task-specific; common examples include accuracy for classification and Dice score for segmentation.
The rejection curve is then defined as a curve consisting of all points $(r, \mathrm{M}(r))$ for $r \in [0, 1]$.

One can interpret rejection curves in the following way: if a method provides good uncertainty estimates, then mistakes made by the model are more likely to be assigned high uncertainty values.
Removal of such samples would then lead to a significant increase of the target metric, and the curve would rise quickly in the beginning.
Better uncertainty estimation methods relate to higher curves.
The best possible curve is theoretically the one where an `oracle' sets uncertainty of $0$ to all correct predictions and some positive value to model errors.

Nevertheless, there are some difficulties in rejection curves pairwise comparison.
% Two rejection curves are hard to compare.
To provide a single value that quantifies uncertainty estimation quality, we calculate the overall area between a rejection curve constructed for the specific method and the best possible rejection (`oracle') curve on an interval $[0, r_{\max}]$ that we further refer to as the \textit{area above the curve} ($\mathrm{AAC}$):
\begin{equation}
    \mathrm{AAC}_{\mathrm{method}} = \int_0^{r_{\max}} \bigl( \mathrm{M}_{\mathrm{oracle}}(r) - \mathrm{M}_{\mathrm{method}}(r) \bigr)dr.
    \label{eq:aac}
\end{equation}
The better the method, the closer its rejection curve is to the `oracle' curve; $\mathrm{AAC}$ of the perfect uncertainty estimator is $0$.
Calculating the area between curves on their entire domain, i.e. $r_{\max} = 1$, leads to decrease in the usefulness of this metric as at higher rejection  rates the importance of each individual error increases.
For real-life applications, however, it is the smaller rejection rates that warrant more attention, since the model that rejects predictions for large fractions of samples (meaning high rejection rate) would be of small use.
For the reasons above we set $r_{\max} = 0.8$.
In practice, the integral~\eqref{eq:aac} is computed numerically.

\subsection*{Experimental evaluation}
% \todo[inline]{TODO.}
We validate our approach across four different machine learning tasks in the field of digital medicine: 
(1) binary image classification in the form of blood agglutination detection on the BloodyWell~\cite{korchagin2024image} dataset;
(2) binary image segmentation in the form of nodule segmentation in lung CT scans of the LID-IDRI~\cite{armato2011lung} dataset;
(3) multi-class image segmentation in the form of optic disc and cup segmentation in retinal fundus images of the RIGA~\cite{almazroa2018retinal} dataset;
and (4) multiple-choice medical question answering on the PubMedQA~\cite{jin2019pubmedqa} dataset.
All mentioned datasets contain multi-expert annotations, a feature that is crucial for the purposes of our paper.

Classification problems (1) and (4) adopt the standard workflow for the evaluation detailed below.
Segmentation tasks (2, 3) are viewed as pixel-level classification problems, which allows for simple adaptation of our approach.
During evaluation, for each rejection rate $r$, a fraction $r$ of pixels with the highest uncertainty values is discarded from each image instead of discarding samples entirely.
This approach is more natural in real-life scenarios, where a rough segmentation map would be more useful than a no segmentation map at all.

We refer to our main approach, estimating aleatoric uncertainty from CAE predictions, as \textit{Total variance, CAE}.
For comparison, several baselines are introduced:
\begin{itemize}
    \item \textit{Single model:} use $(1 - \mathrm{maximum\, confidence})$ predicted by the model for each sample;
    \item \textit{Prediction variance:} evaluate prediction variance of the base ensemble via~ \eqref{eq:epistemic};
    \item \textit{Total variance, base:} compute total variance estimates from solely the base ensemble predictions. Here both epistemic and aleatoric uncertainty are estimated using predictions from the base ensemble;
    \item \textit{Total variance, experts:} compute a total variance estimation where expert annotations are used directly in place of a CAE.
\end{itemize}

The first three baselines are weaker versions of our approach that lack certain components of it.
The last one \textit{(Total variance, experts)} would be infeasible in practice, as answers from multiple experts are not typically available during inference.
However, it showcases usefulness of our approach by providing additional evidence of the expert-based confidence labels.

A plethora of uncertainty estimation techniques widespread in the literature are also evaluated to further solidify the effectiveness of the proposed solution.
This includes pseudo-ensembling techniques such as MCMC~\cite{welling2011bayesian} and MC Dropout~\cite{gal2016dropout}; an entropy-based uncertainty evaluation approach~\cite{gal2016uncertainty}; test-time augmentations~\cite{wang2018test, tonolini2024bayesian}; and state-of-the-art methods such as Hybrid Uncertainty Quantification~\cite{vazhentsev2023hybrid} (HUQ) and Adaptive Bayesian Neural Networks~\cite{franchi2024make} (ABNN). 

For classification tasks, an additional measure of uncertainty estimation quality is provided.
Uncertainty estimation can be viewed as an error detection task, where models responses are classified into correct and erroneous predictions, with higher uncertainty values corresponding to higher probabilities of error in predicted classes.
% This aligns greatly with the definition of uncertainty.
The quality of such prediction classification corresponds to the quality of uncertainty estimation.
We choose area under the receiver-operator curve (AUROC) as a single-value measure of classification quality since it is independent of classification threshold choice and insensitive to class imbalance~\cite{fawcett2006introduction}.

\section*{Results}

This section presents an extensive empirical analysis of the proposed solution.
First, our solution is compared to various baselines and state-of-the-art methods as described in the previous section.
Results show that methods that use expert annotations outperform all the other approaches.
Next, an analysis of various aspects of our approach is presented, including potential ways of improving efficiency.

\subsection*{Experts provide good aleatoric uncertainty estimates}

\begin{figure*}[h!]
    \centering
    \begin{subfigure}[b]{0.5\textwidth}
        \centering
        \includegraphics[height=2.4in]{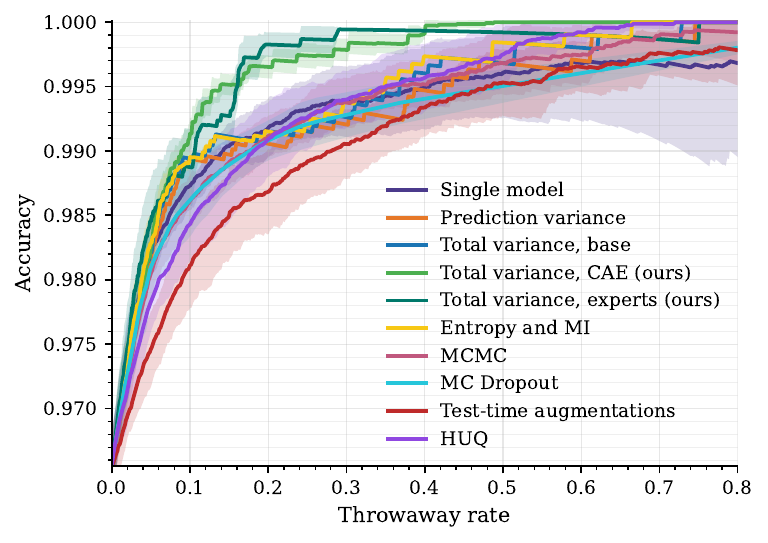}
        \caption{Image classification (BloodyWell)}
    \end{subfigure}%
    ~ 
    \begin{subfigure}[b]{0.5\textwidth}
        \centering
        \includegraphics[height=2.4in]{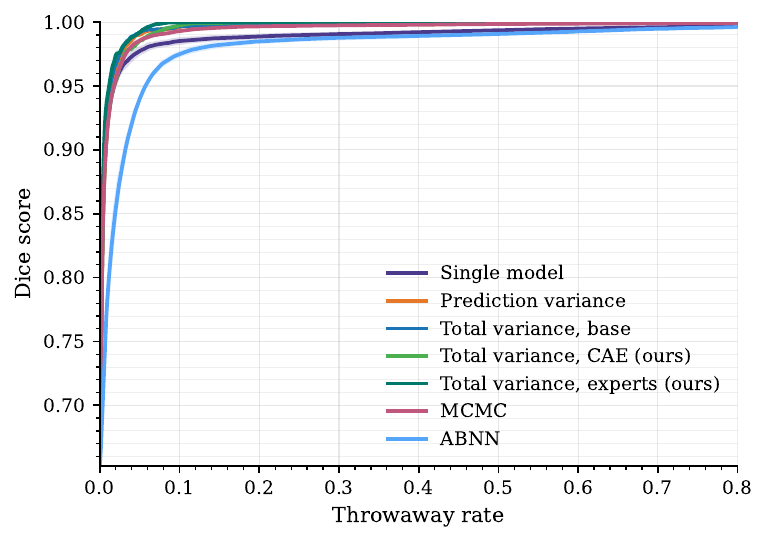}
        \caption{Binary segmentation (LIDC-IDRI)}
        \label{subfig:gr-lidc}
    \end{subfigure}
    \centering
    \begin{subfigure}[b]{0.5\textwidth}
        \centering
        \includegraphics[height=2.4in]{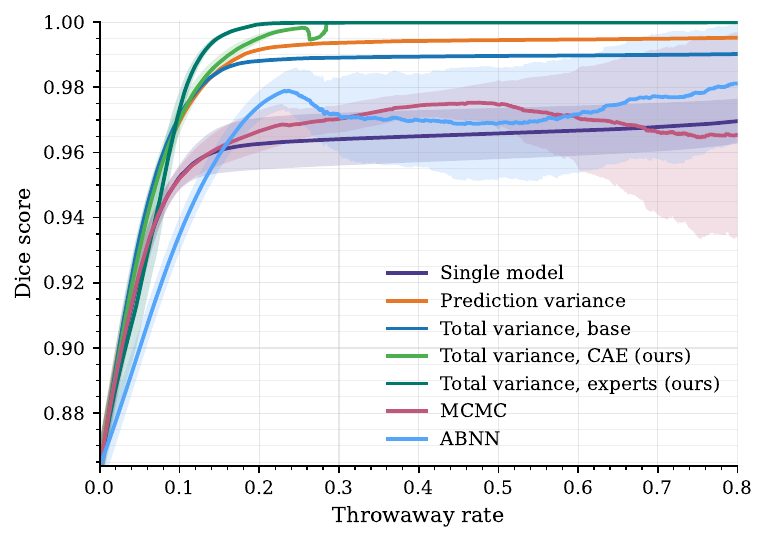}
        \caption{Multiclass segmentation (RIGA)}
        \label{subfig:gr-riga}
    \end{subfigure}%
    ~ 
    \begin{subfigure}[b]{0.5\textwidth}
        \centering
        \includegraphics[height=2.4in]{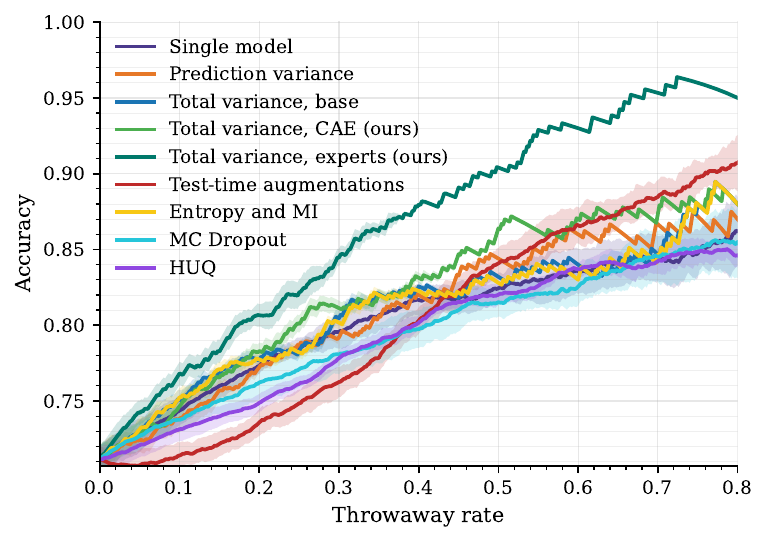}
        \caption{Multiple choice QA (PubMedQA)}
    \end{subfigure}
    \caption{Rejection curves for proposed methods across four different machine learning tasks. Shaded areas represent the standard deviation between ensemble models. The higher the curve, the better.}
    \label{fig:global-results}
\end{figure*}

Table~\ref{tab:global-results} presents the AAC values for the proposed methods across four machine learning tasks: binary image classification, binary and multiclass segmentation, and multiple-choice question answering. 
Table~\ref{tab:auroc-results} contains AUROC values for error detection of uncertainty estimation methods in binary image classification and multiple-choice question answering.
The detailed rejection curves are presented in Figures~\ref{fig:global-results}.

\begin{table}[htbp]
\centering
\small
\begin{tabular}{l c c c c c}
\toprule
\multirow{2}{*}{UE Method} & \multirow{2}{*}{Uses experts} & \multicolumn{4}{c}{AAC, $\times 10^{-3}$, $\downarrow$} \\
\cmidrule{3-6} 
{} & {} & Image class. & Binary seg. & Multiclass seg. & Multiple choice QA \\
% UE Method & Uses experts & Image cls. & Binary seg. & Multiclass seg. & Multiple choice QA \\
\midrule
Single model & & $5.11$ & $8.27$ & $30.07$ & $116.56$ \\
\midrule
Prediction variance & & $4.89$~$(\cgreen{0.043}{-4.3\%})$ & $2.93$~$(\cgreen{0.646}{-64.6\%})$ & $9.72$~$(\cgreen{0.677}{-67.7\%})$ & $110.93$~$(\cgreen{0.048}{-4.8\%})$ \\

Total variance, base & & $4.28$~$(\cgreen{0.02}{-16.2\%})$ & $2.63$~$(\cgreen{.682}{-68.2\%})$ & $12.54$~$(\cgreen{0.583}{-58.3\%})$ & $110.32$~$(\cgreen{0.054}{-5.4\%})$ \\

Total variance, CAE & \checkmark, training & $2.18$~$(\cgreen{0.573}{-57.3\%})$ & $2.44$~$(\cgreen{0.705}{-70.5\%})$ & $6.45$~$(\cgreen{0.786}{-78.6\%})$ & $100.85$~$(\cgreen{0.135}{-13.5\%})$ \\

Total variance, experts & \checkmark, inference & $2.37$~$(\cgreen{0.536}{-53.6\%})$ & $1.38$~$(\cgreen{0.833}{-83.3\%})$ & $6.27$~$(\cgreen{0.791}{-79.1\%})$ & $66.11$~$(\cgreen{0.433}{-43.3\%})$ \\
\midrule

Entropy and MI~\cite{gal2016uncertainty}& & $4.00$~$(\cgreen{0.217}{-21.7\%})$ & -- & -- & $111.05$~$(\cgreen{0.047}{-4.7\%})$ \\

MC Dropout~\cite{gal2016dropout} & & $5.67$~$(\cred{0.109}{+10.9\%})$ & -- & -- & $123.12$~$(\cred{0.056}{+5.6\%})$ \\

MCMC~\cite{welling2011bayesian} & & $4.86$~$(\cgreen{0.049}{-4.9\%})$ & $3.96$~$(\cgreen{0.521}{-52.1\%})$ & $27.12$~$(\cgreen{0.098}{-9.8\%})$ & -- \\

Test-time augmentations~\cite{wang2018test, tonolini2024bayesian} & & $6.96$~$(\cred{0.362}{+36.2\%})$ & -- & -- & $117.48$~$(\cred{0.008}{+0.8\%})$ \\

HUQ~\cite{vazhentsev2023hybrid} & & $4.68$~$(\cgreen{0.084}{-8.4\%})$ & -- & -- & $124.35$~$(\cred{0.067}{+6.7\%})$ \\

ABNN~\cite{franchi2024make} & & -- & $15.53$~$(\cred{0.878}{+87.8\%})$ & $27.74$~$(\cgreen{0.077}{-7.7\%})$ & -- \\
\bottomrule
\end{tabular}
\caption{\label{tab:global-results} Comparison of uncertainty estimation methods using rejection curves.
AAC stands for the Area Above Curve and is normalized with respect to the optimal rejection-accuracy curve for a given model.
The value in the bracket shows improvement over the baseline single-model method.}
\end{table}

Across all considered machine learning tasks, the proposed approach \emph{Total variance, experts} outperforms other considered solutions and provides improvements of $13\%$ to $78\%$ over outputs generated by a single model.
Using real expert annotations during inference improves AAC values by an additional $2\%$ to $43\%$. 
Thus, expert-annotated uncertainty during either training or inference provides a significant quality boost compared to the case without such knowledge.
In three out of four tasks, using total variance rather than prediction variance improves uncertainty estimation, motivating its further usage.
We discuss the specifics for the separate tasks below.

\subsubsection*{Image classification}

% Image classification is generally considered an easier task compared to image segmentation, which is confirmed by the high accuracy of $96.55\% \pm 0.55\%$ on the BloodyWell dataset.

The BloodyWell dataset~\cite{korchagin2024image} contains the most comprehensive expert annotations in this study, with each sample labeled by six independent voters.
The Total Variance method leveraging these expert annotations outperforms the Total Variance with the base ensembles for both epistemic and aleatoric component approaches by $44.6\%$. 
The total variance method using a CAE to approximate experts provides even larger improvement in AAC over the base ensemble: $49.1\%$.
% Notably, the performance of the CAE-based method was nearly equivalent to that of the method that uses experts' markup during training, with only a $1.8\%$ difference in AAC. 
Notably, the CAE-based method outperforms the method that uses experts' markup during training with a $8\%$ difference in AAC. 
We suggest the reason for it is that the CAE's, neural networks', outputs estimate uncertainty of the neural net ensemble even better than human votes.
Furthermore, the outputs from CAE have more diversity than experts' votes since each expert actually chose one of the five answers.
It is also interesting that rejection curves for the total variance methods with both a CAE and experts' markup demonstrate the most sharp incline.
It highlights the practical utility of these approaches: rejection of smaller data part provides the same increase in accuracy compared to other methods.
It is also significant to mention that the total variance with a CAE reaches perfect accuracy extremely faster than all the other methods (less than $40\%$ of data have to be rejected). 

Among the baseline methods, the highest performance was achieved by the approach that employs entropy for aleatoric uncertainty estimation and mutual information (MI) for epistemic one~\cite{gal2016uncertainty}. 
It demonstrates an approximately $22\%$ improvement over the single model method.
Single net baseline is also slightly outperformed by HUQ~\cite{vazhentsev2023hybrid} and MCMC~\cite{welling2011bayesian} approaches, by $8\%$ and $5\%$ respectively. 
Test-time augmentations (TTA) demonstrated the lowest AUC.  
% MC Dropout and MCMC provide similar quality, as expected, since they share a similar general idea (a Bayesian approach and Monte Carlo sampling).

% We did not implement ABNN for the BloodyWell because it is not designed for classification and does not work well for it.

\begin{table}[tbp]
\centering
\small
\begin{tabular}{l c c c}
\toprule
\multirow{2}{*}{UE Method} & \multirow{2}{*}{\hspace{1em}Uses experts\hspace{1em}} & \multicolumn{2}{c}{AUROC, $\uparrow$} \\
\cmidrule{3-4} 
{} & {} & \hspace{1em}Image classification\hspace{1em} & \hspace{1em}Multiple choice QA\hspace{1em} \\
% UE Method & Uses experts & Image cls. & Binary seg. & Multiclass seg. & Multiple choice QA \\
\midrule
Single model & & $0.889 \pm 0.033$ & $0.693 \pm 0.022$ \\
\midrule
Prediction variance & & $0.888 \pm 0.008$ & $0.702 \pm 0.011$ \\

Total variance, base & & \underline{$0.900 \pm 0.008$} & $0.709 \pm 0.012$ \\

Total variance, CAE & \checkmark, training & $\mathbf{0.942 \pm 0.008}$ & \underline{$0.732 \pm 0.014$} \\

Total variance, experts & \checkmark, inference & $\mathbf{0.942 \pm 0.008}$ & $\mathbf{0.818 \pm 0.007}$ \\
\midrule

Entropy and MI~\cite{gal2016uncertainty}& & $0.879 \pm 0.009$ & $0.707 \pm 0.012$ \\

MC Dropout~\cite{gal2016dropout} & & $0.852 \pm 0.025$ & $0.671 \pm 0.022$ \\

MCMC~\cite{welling2011bayesian} & & $0.887 \pm 0.020$ & -- \\

Test-time augmentations~\cite{wang2018test, tonolini2024bayesian} & & $0.842 \pm 0.033$ & $0.670 \pm 0.013$ \\

HUQ~\cite{vazhentsev2023hybrid} & & $0.887 \pm 0.023$ & $0.662 \pm 0.010$ \\
\bottomrule
\end{tabular}
\caption{\label{tab:auroc-results} Comparison of uncertainty estimation methods in error detection, where predicted uncertainty score is used as confidence in classification of model responses into errors and correct answers.
AUROC stands for the Area Under the Receiver-Operator Curve.
The best solution for each task is in bold and the second best solution is underlined.}
\end{table}

\subsubsection*{Binary and multiclass segmentation}

Results for segmentation tasks are mostly similar, with ensemble-based solutions offering at least a $58\%$ improvement over the baseline.
Notably, multiclass segmentation is the only task where total variance is outperformed by prediction variance with a $29\%$ better AAC value.
An interesting observation is that in Fig.~\ref{subfig:gr-riga}, variance-based approaches show faster Dice improvement than \textit{Total variance, experts} up until around a rejection rate of $r = 0.1$, which suggests that, in certain applications, machine learning models can better highlight the most probable error cases.

Potential utility of pseudo-ensembling approaches for segmentation tasks is showcased by MCMC providing better uncertainty estimates than single models with up to a $52\%$ improvement.
However, at higher rejection rates MCMC predictions become unstable, leading to highly varied performance, as can be seen in Fig.~\ref{subfig:gr-riga}.

The most prominent and somewhat surprising difference is that ABNN performs on par with MCMC for multiclass segmentation but struggles in binary segmentation, resulting in an almost $2\times$ higher AAC.
Fig.~\ref{subfig:gr-lidc} reveals a slow initial improvement in Dice when using ABNN uncertainty estimates, which contributed greatly to a big AAC.
Even at higher rejection rates, the performance of ABNN barely surpasses that of a single model.
For multiclass segmentation, ABNN provides inconsistent uncertainty estimates, similar to MCMC (see Fig.~\ref{subfig:gr-riga}).

\subsubsection*{Multiple choice question answering}

Ensemble models achieve an accuracy of $71.18\% \pm 0.85\%$ on the PubMedQA dataset~\cite{jin2019pubmedqa}.
Results showcase that even a small number of expert annotators (two) provides enough information to improve aleatoric uncertainty estimation.
% , and at throwaway rate $t = 0.2$ (every fifth sample is not classified) accuracy improves to $80.62\%$. 
Overall, expert annotations allow for a $43\%$ improvement in AAC over the single-model baseline. 
The expert-aware ensemble also outperforms the standard models by a modest $14\%$ but does not reach the uncertainty estimation quality of real experts.
This can be attributed to multiple factors, mainly a rather small training size of only 450 samples, as well as a general lack of variety in expert responses.
It also may happen due to the fact that the accuracy of the votes' prediction is only $72\%$.
Complete experts' consistency is achieved for $69\%$ of test data.
It is also interesting that for $16\%$ of test data the experts gave the opposite answers (one ``yes'', the other ``no'').
Such samples can be especially difficult to answer.

The best baseline approaches with almost the same AAC are ensemble variance and, as for the BloodyWell dataset, entropy. 
The lowest quality is showed by the HUQ method. 
In contrast to BloodyWell, test-time augmentations~\cite{tonolini2024bayesian} performed almost as well as single net method. 
\subsection*{How many experts are needed for accurate uncertainty estimates?}
Assessments from multiple experts in practice are an expensive resource.
Moreover, experts vary greatly in their accuracy~\cite{almazroa2017agreement}, which might impact the quality of a system relying on their repsonses.
% Moreover, it is extremely important to find highly qualified specialists because the quality of their work affects the overall system's quality.
This necessitates a study on how the quality of the proposed methods depends on the number of experts used.

\begin{figure*}[t]
    \centering
    \begin{subfigure}[b]{0.49\textwidth}
        \flushleft
        \includegraphics[width=\linewidth]{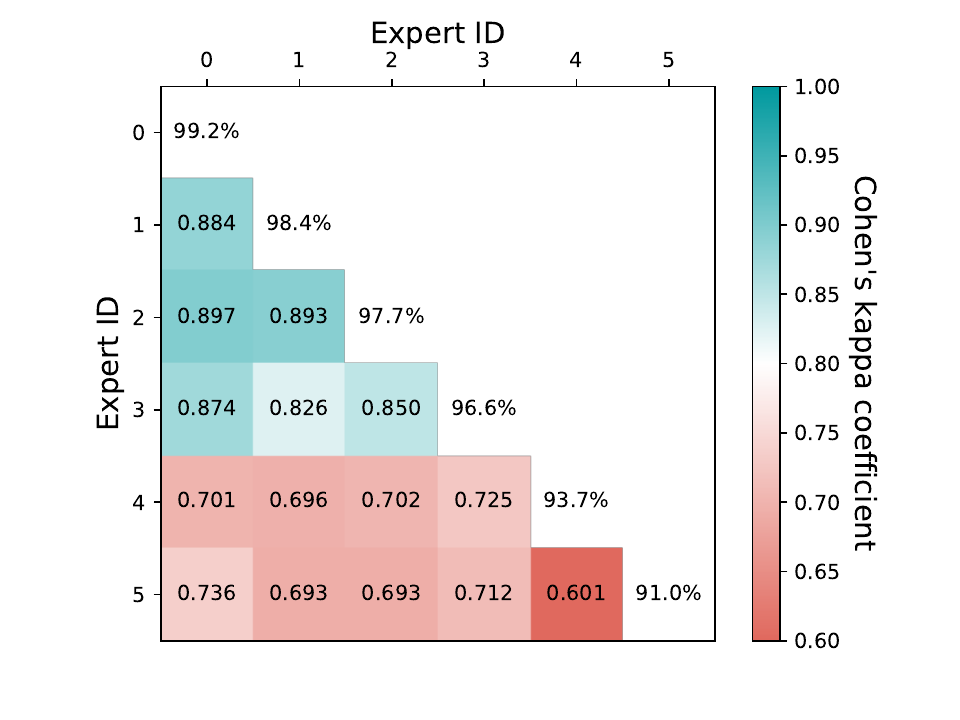}
        \caption{}
        \label{fig:cohen}
    \end{subfigure}
    ~
    \begin{subfigure}[b]{0.49\textwidth}
        \centering
        \includegraphics[width=\linewidth]{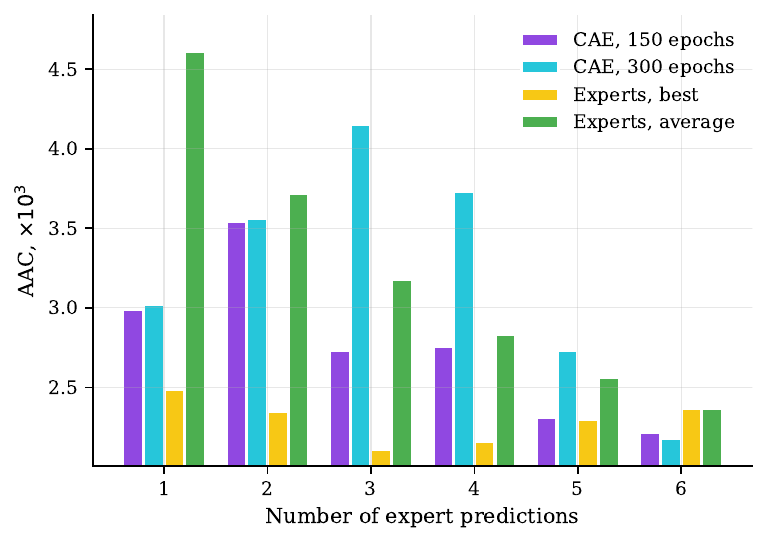}
        \caption{}
        \label{fig:expert-size}
    \end{subfigure}
    \caption{\textbf{(a)} The pairwise consistency between experts estimated as the Cohen's kappa coefficient. Values on the diagonal are experts' accuracies. \textbf{(b)} Uncertainty estimation quality by the law of total variance where aleatoric uncertainty is estimated from different sources. \textit{Experts, best} stands for using the best $k$ experts based on accuracy, and \textit{Experts, average} is computed by averaging AAC for all possible sets of $k$ experts.}
\end{figure*}

The BloodyWell dataset was chosen for this experiment since it contains the most extensive expert markup.
In order to rank experts, we calculated their quality on the entire BloodyWell dataset using binary ground truth values. 
If an expert gave an uncertain correct answer, it counted towards final accuracy with a 0.75 coefficient.
Samples where an expert abstained from an answer were simply discarded.
Accuracies of experts, as well as consistency between their responses estimated as Cohen's kappa coefficient~\cite{cohen1960coefficient}, are presented in Figure~\ref{fig:cohen}.

We compare uncertainty estimation quality when using different sizes of experts groups for either direct aleatoric uncertainty estimation via \eqref{eq:aleatoric} or training a CAE.
For the first case, both the quality of $k$ best experts (denoted \textit{Experts, best}) and the average quality of all possible subgroups of $k$ experts (denoted \textit{Experts, average}) were computed for each $k \in \{1, \ldots, 6\}$.
CAEs were only trained on the groups of $k$ best experts as training an ensemble for each expert subset is computationally infeasible.
Two different training regimes were additionally considered, as it was theorized that ensembles trained on smaller group sizes would benefit from shorter training time -- 150 epochs instead of 300 epochs used previously.
Results are presented in Figure~\ref{fig:expert-size}. 

% \begin{table}[htbp]
% \centering
% \small
% \begin{tabular}{l c c c c c c}
% \toprule
% \multirow{2}{*}{UE Method} & \multicolumn{6}{c}{Number of experts used} \\
% \cmidrule{2-7} 
% {} & 1 & 2 & 3 & 4 & 5 & 6 \\
% \midrule
% Total variance, CAE, 700ep & 3.02 & 3.56 & 4.15 & 3.73 & 2.73 & 2.18 \\
% Total variance, CAE, 550ep & 2.99 & 3.54 & 2.73 & 2.76 & 2.31 & 2.22 \\
% Total variance, experts best & 2.49 & 2.35 & 2.11 & 2.16 & 2.30 & 2.37 \\
% Total variance, experts average & 4.61 & 3.72 & 3.18 & 2.83 & 2.56 & 2.37 \\
% \bottomrule
% \end{tabular}
% \caption{\label{tab:experts_number} AAC, $\times 10^{-3}$, $\downarrow$, for Total Variance method with CAE and experts' markup for different number of experts for BloodyWell dataset.}
% \end{table}

The average quality of an expert group grows with increase in group size, with six experts outperforming an average individual expert by a factor of almost 2.
This is not the case, however, for when the groups of best experts are considered.
Six best specialists provide predictions only marginally better than the best expert, and the best group size, somewhat surprisingly, is three.
This highlights the importance of ``quality over quantity'' approach for data acquisition, especially for the scenario at hand.
Another thing to note is that the accuracy of an expert correlates strongly with the performance of the uncertainty estimates computed from their predictions, with experts achieving AAC values of 2.49, 3.93, 4.69, 4.86, 5.03 and 6.64 respectively for expert ID 0 through 5.

The CAE trained for 150 epochs mostly improves when trained on bigger expert groups, whereas the CAE trained for 300 epochs suffers from poor uncertainty estimation quality and performs comparatively worse unless all six experts are used.
This signals that each training scenario requires careful hyperparameter tuning to achieve the best possible outcomes.
Additionally, a CAE trained on the best experts outperforms direct evaluation by average specialists, which further solidifies the need for precision and quality in expert labeling.

% Notably, one and two best experts provide higher quality than three and four ones. 
% The probable reason is that the top-2 experts have really great quality.
% In order to make a verification, the results are compared to the performance of the Total Variance method that uses experts' markup with the same experts combinations. 

% % % % %
% For all experts' combinations, except for the number of experts equal to two, the best performance was provided by the combination of top-quality experts. 
\subsection*{Relation of uncertainty estimation quality to ensemble size}

\begin{figure}[t]
    \centering
    \includegraphics[height=3.0in]{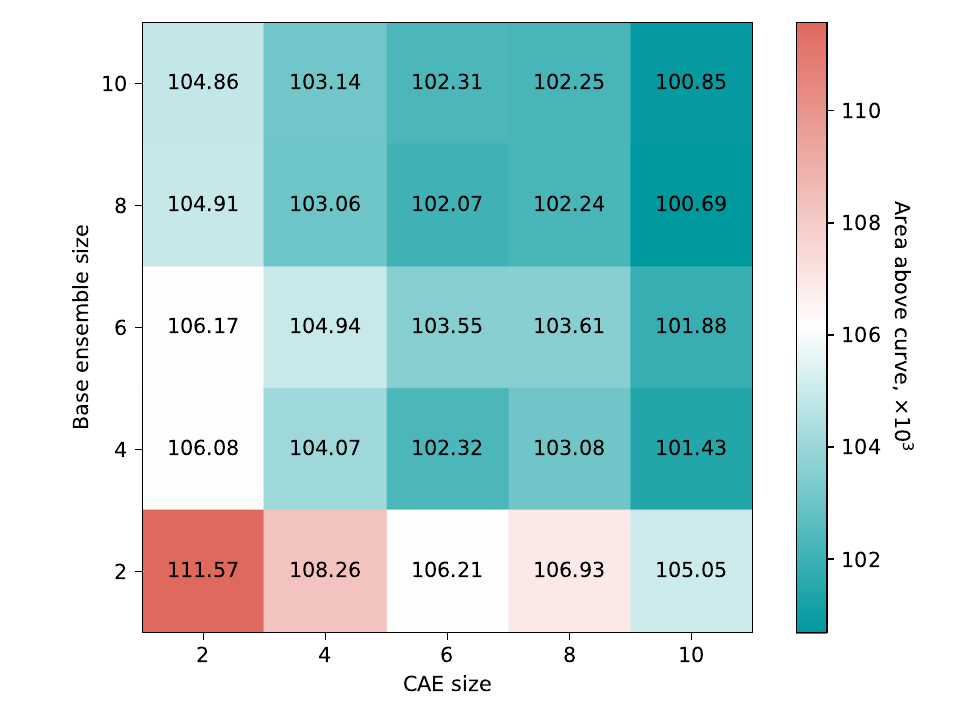}
    \caption{Relation of uncertainty estimation quality AAC on the multiple choice question answering task to the sizes of the base ensemble and the CAE. Lower values are better.}
    \label{fig:ensemble-size}
\end{figure}

In this experiment, the sizes of both the base ensemble and the CAE were varied to assess how they affect uncertainty estimation.
Figure~\ref{fig:ensemble-size} presents AACs achieved by different combinations of ensemble sizes on the multiple-choice question answering task.

In general, uncertainty estimation improves with increasing ensemble size.
The estimate of epistemic uncertainty saturates faster than that of aleatoric uncertainty, with a four-model base ensemble generally performing on par with a ten-model ensemble.
On the contrary, the performance of a CAE steadily increases with greater ensemble sizes up to ten models.
We also see that, with a base ensemble of size 4 and a CAE of size 4, the proposed \textit{Total variance, CAE} approach outperforms a ten-model \textit{Total variance, base} approach. 
\subsection*{Perfomance of proposed solutions in real-life scenarios: a simulation}
In practice, we suggest to use our method in the following way: the most uncertain part of the data is rejected, the rest of the samples are classified by the second-opinion system.
Then the rejected data is classified by the human expert with some accuracy. 
After that the overall accuracy of the test increases.

We suppose the expert can select the samples that they classify with a $100\%$ accuracy. 
For the rest of samples the test can be repeated by the human, and again they choose the part of data where they are sure in the result.
The accuracy increases again.
Then we get the following formula of achieved accuracy after $n$ repetitions of the test:
\begin{equation}
(1 - r) \cdot a_{cl} + \frac{r \cdot a_{exp} \cdot (1 - (1 - a_{exp})^{n - 1})}{a_{exp}} = (1 - r) \cdot a_{cl} + r \cdot (1 - (1 - a_{exp})^{n - 1}),
   \label{eq:accuracy_after_repetions}
\end{equation}
where $a_{exp}$ is the accuracy of the expert, $a_{cl}$ is the accuracy of the classifier, $r$ is the rejected data part.
The second summand in the equation~\ref{eq:accuracy_after_repetions} is the sum of the geometric progression with the initial value $r \cdot a_{exp}$ and ratio $1 - a_{exp}$.
Here we suppose that the expert classifies perfectly and surely $95.6\%$ of the data since this is the mean accuracy of the experts on the BloodyWell dataset.

\begin{figure}[ht]
    \centering
    \includegraphics[height=3in]{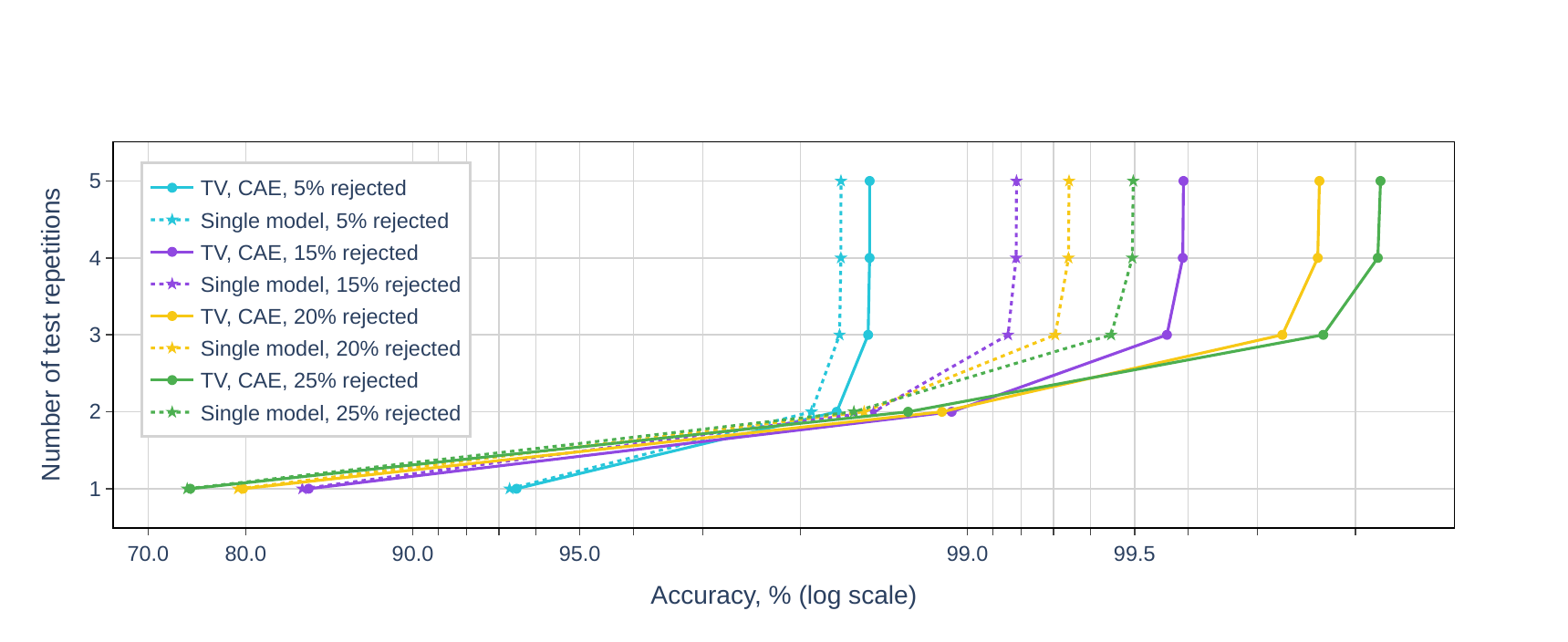}
    \caption{Accuracy that can be achieved with the Total Variance with CAE and a single model as the uncertainty estimation method for different rejection rates.}
    \label{fig:paretto}
\end{figure}

Figure~\ref{fig:paretto} illustrates the accuracy that can be achieved with the Total Variance with CAE and a single model as the uncertainty estimation method for different rejection rates for a different number of test repetitions.

% TODO: include discussion

\subsection*{Weighting coefficient}
One possible extension of the proposed approach is to calculate total uncertainty as a weighted sum of the epistemic and aleatoric components, as in
\begin{equation}
    \mathrm{TU}(y) = 2\big[ \alpha \, \mathrm{EU}(y) + (1 - \alpha) \mathrm{AU}(y) \big],
\end{equation}
for $\alpha \in [0, 1]$, with this equation simplifying to~\eqref{eq:total-variance} at $\alpha = 0.5$. AAC values for different values of $\alpha$ across various tasks are presented in Table~\ref{tab:TV_coef}.

\begin{table}[htbp]
\centering
\small
\begin{tabular}{l c c c c c c c c c c c}
\toprule
\multirow{2}{*}{UE Method} & \multicolumn{11}{c}{Weighting coefficient, $\alpha$} \\
\cmidrule{2-12} 
{} & 0.0 & 0.1 & 0.2 & 0.3 & 0.4 & 0.5 & 0.6 & 0.7 & 0.8 & 0.9 & 1.0 \\
\midrule
Image classification & 2.58 & 2.47 & 2.38 & 2.30 & 2.23 & 2.18 & 2.15 & 2.14 & 2.17 & 2.24 & 4.89 \\
Image segmentation & 19.21 & 8.09 & 7.55 & 7.10 & 6.73 & 6.45 & 6.24 & 6.10 & 6.04 & 6.06 & 9.72 \\
Multiple-choice QA & 103.12 & 103.00 & 102.20 & 101.75 & 101.33 & 100.85 & 100.67 & 100.58 & 101.05 & 102.46 & 111.27 \\
\bottomrule
\end{tabular}
\caption{\label{tab:TV_coef} Dependency of AAC, $\times 10^{-3}$, on the weighting coefficient of epistemic uncertainty component for the Total Variance approach with CAE.}
\end{table}

Results suggest that an optimal value of $\alpha$ is greater than 0.5 and lays somewhere in the range of $[0.6, 0.8]$.
This value, however, is dependent on the task or the dataset and is not supported by theoretical evidence.
The difference in quality of uncertainty estimation between the proposed $\alpha = 0.5$ and the optimal value is mostly negligent, being at most $7\%$ for segmentation and $2\%$ for classification. 
Another important  observation is that across all tasks, the AAC for $\alpha=0.9$ is up to 2 times better than the AAC for $\alpha=1.0$. 
It means that even a small addition of expert knowledge dramatically improves quality, justifying our overall approach once again. 
% The best coefficient was $0.7$, but its difference from $0.5$ is not significant.
In fact, all values of $\alpha \in (0, 1)$ for both datasets yield comparable AAC values; therefore, we use $\alpha=0.5$ and suggest this value in practice.
\subsection*{One-ensemble mixed loss approach}

The two-ensemble approach is ineffective in practice, as it requires significant time and memory for both training and inference.  
Therefore, we suggest a single-ensemble method. 
The idea is to use expert votes and hard labels simultaneously during training: some samples have soft ground truth, while others have hard labels.
The loss function is mixed: it is calculated separately for soft and hard samples, and the overall loss is a weighted sum with the weighting hyperparameter $
\alpha$:
\begin{equation}
    \mathrm{Loss}_{\mathrm{total}} = \alpha  \cdot \mathrm{Loss}_{\mathrm{hard}} + (1 - \alpha)  \cdot \mathrm{Loss}_{\mathrm{soft}}.
    \label{eq:mixed_loss}
\end{equation}

% It has slightly lower quality but requires twice as few neural network models.

The motivation behind separating losses is flexibility. 
Sometimes expert votes are much more precise, and 
the accuracy of hard and soft labels can differ significantly. 
Therefore, one may want to prioritize one type of label while considering an alternative source of information.

The obtained ensemble is used to estimate both epistemic and aleatoric uncertainty using equations~\ref {eq:epistemic} and~\ref{eq:aleatoric}, respectively.

\begin{figure}[ht]
    \centering
    \includegraphics[height=2.4in]{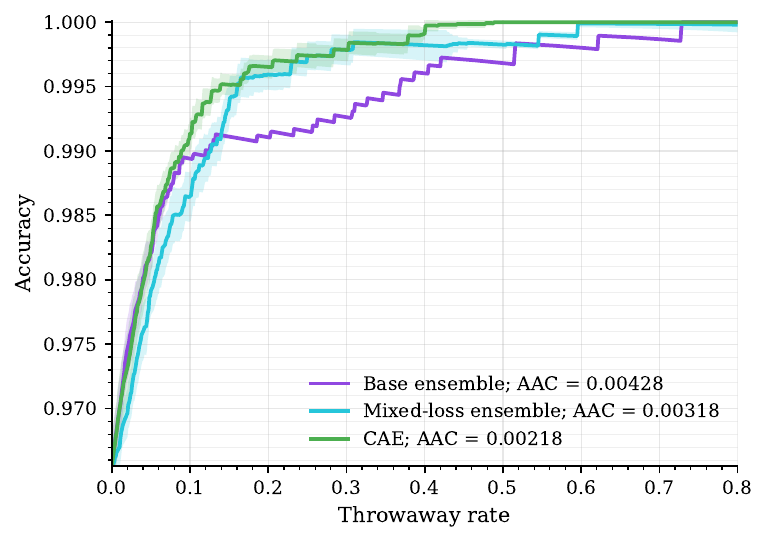}
    \caption{Accuracy-rejection curves compared between total variance methods using a base ensemble, an ensemble trained with a mixed loss, and a CAE on the BloodyWell dataset.}
    \label{fig:mixed-loss}
\end{figure}

We validate this idea using the BloodyWell dataset, which has the most extensive expert annotations.
The corresponding accuracy-rejection curves are demonstrated in Fig.~\ref{fig:mixed-loss}.

The achieved AAC is $3.18 \times 10^{-3}$, representing a $37.7\%$ improvement over the single-model approach.
The mixed-ensemble method reduces the error rate by $8.4$ times, from $3.43\%$ to $0.41\%$, by rejecting 20\% of the data.
Nevertheless, this performance remains lower than that of the approach employing separate ensembles for epistemic and aleatoric uncertainty estimation. 
This result supports the hypothesis that expert annotations are more useful for estimating aleatoric components, whereas epistemic uncertainty is better characterized by categorical data. 
It is also worth mentioning that the mixed-loss ensemble approach significantly outperforms all the baseline methods.
\subsection*{Leveraging LLMs as ``oracles'' to substitute human experts}

% \todo[inline]{Add more explicit formulation of LLMs. Why this task is unique. }

Most of the time, annotations from different experts are unavailable or expensive to obtain.
As shown above, it is not always possible to effectively learn patterns in expert annotations.
On the other hand, recent advancements in large language model development allowed for the creation of models that possess knowledge on a wide variety of topics without the need for any fine-tuning.
This leads to the idea of using such an LLM as an ``oracle'' to provide expert-level annotations in the absence of human experts.

We validate this idea by prompting a large language model to provide a response to each question in the PubMedQA dataset and evaluating its confidence in the response (see Methods for specifics).
For each question, multiple answers are generated and treated as separate expert responses.
We then use the provided responses to estimate aleatoric uncertainty.

\begin{figure}[ht]
    \centering
    \includegraphics[height=2.4in]{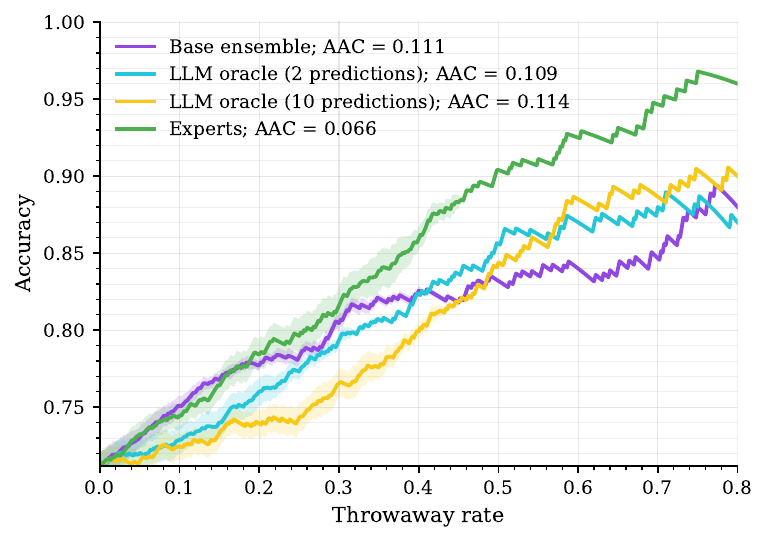}
    \caption{Accuracy-rejection curves compared between total variance methods using a base ensemble, an LLM as an oracle and experts on the PubMedQA dataset.}
    \label{fig:oracles}
\end{figure}

% This method is denoted $\mathrm{TV}(\mathrm{Ens}, \mathrm{Orac})$, and accuracy-rejection curves are presented in Fig~\ref{fig:oracles}.
This method is denoted \textit{LLM oracle}, and accuracy-rejection curves are presented in Fig~\ref{fig:oracles}.

The proposed method generally performs the same or worse at lower rejection rates, but at higher values, uncertainty estimation improves significantly.
When more predictions are generated, performance improves at higher rejection rates but degrades for smaller values.
Metric-wise, \textit{LLM oracle} outperforms \textit{Total variance, base} with AAC value of $0.109$.
It should be noted, however, that in practical scenarios better performance at lower rejection rates would be more important.
Still, this result shows potential for future use of LLMs as a source of expert-level annotations.

\section*{Discussion}

% \todo[inline]{TODO by Aleksei Kh. Include specific results, numbers, details on experts - whatever you found valuable}

Reliable uncertainty estimates are essential for the adoption of AI systems in medical decision-making, yet current approaches often fall short of clinical requirements. 
In this work, we show that incorporating expert knowledge directly into the uncertainty estimation process leads to more trustworthy confidence. 
Our method uses a variance-based decomposition of total predictive uncertainty into epistemic and aleatoric components, explicitly modeling the aleatoric part with a model that mimics expert-informed soft labels.
The nature of these confidence scores expressed by human experts and AI models are close, allowing a similar method to take them into account during model training or inference. 
As a result, our principled ensemble-based uncertainty estimates, including the CAE approach, provide superior uncertainty evaluation across all examined tasks.

The empirical study demonstrates that experts provide information about data complexity and ambiguity that is not recoverable from hard labels alone, which are proven to be insufficient to train a powerful uncertainty evaluation model. 
Models trained with expert soft labels are better able to flag cases that are genuinely difficult, thereby supporting more effective abstention strategies. These benefits arise even with a minor number of soft labels, suggesting that the approach is compatible with the constraints of clinical workflows. 

The observed gains are consistent across clinically relevant tasks, including multiple-choice question answering, image classification, and segmentation. 
Our CAE method more reliably identifies samples where the model is likely to make mistakes, as confirmed by approximately twofold improvements in AAC. The largest improvements are observed in the segmentation task. 
Furthermore, we introduce a single-ensemble surrogate that retains most of the improvement while substantially lowering computational cost, expanding the scenarios of the model usage.
At the same time, there exists a gap between model-based and expert-based uncertainty at the inference time. The difference is small for high-quality models and aligned expert labels, but for tasks with lower baseline accuracy, such as PubMedQA and LIDC, this gap is larger. 
These findings indicate that improvements in core predictive performance and expert labeling practices will further enhance the effectiveness of expert-guided uncertainty estimation.

Despite various considered optimizations and sensitivity studies, the proposed approach remains computationally expensive as it involves training an entire ensemble of models to ensure the high quality of predictions --- similarly to typical uncertainty estimation methods.
Single-model or low-parameter methods will always surpass ensemble solutions in terms of efficiency.
Whether expert knowledge can be integrated into a more efficient paradigm remains an open question.

Analysis of the correlation between the accuracy of expert labels and the quality of uncertainty estimation reveals that the need to collect precise responses from experienced professionals greatly outweighs the importance of acquiring annotations from large groups of physicians.
Even a small group of top-performing specialists can provide labels of quality high enough to effectively be used in both direct uncertainty estimation as well as training confidence-aware ensembles.
Training appears to be highly dependent on hyperparameter selection, and each scenario requires preliminary hyperparameter tuning to achieve optimal results.

Demonstrating the benefits of utilizing expert confidence in uncertainty estimation, our study gives rise to further research directions, based on the conducted open, benchmark-style evaluation across four datasets. 
One is the active learning strategy~\cite{budd2021survey} of collecting confidences, in which experts label ambiguous or clinically critical cases rather than random ones, thereby reducing the markup costs. 
Another involves the design of mechanisms for collecting confidence signals, including LLM-based scores, into a unified framework~\cite{wei2024measuring}. Finally, moving beyond retrospective evaluation, prospective clinical studies are needed to quantify how improved uncertainty estimates should inform decisions in medical AI, such as avoiding unnecessary follow-up examinations and balancing human-AI workload for improved safety~\cite{kurmukov2024impact,wolff2020economic}.

Taken together, our results support uncertainty modeling based on expert confidence as a practical tool toward more reliable medical AI. 
The experts correctly identify aleatoric uncertainty arising from a sample's internal complexity.
Given a signal from them, the proposed approach yields uncertainty estimates that better reflect data ambiguity. This, in turn, brings AI systems closer to the level of reliability required for integration into real-world clinical decision-making.

\section*{Data availability statement}

All data used in this study are publicly available from open-source repositories on the internet.
Specific sources are provided in the Methods section of the paper and in Supplementary Methods section.

\section*{Funding}

Authors A.K., E.Z. and S.K. were supported by the State Assignment of the Ministry of Science and Higher Education of the Russian Federation (agreement No. FFNU-2025-0045); author E.E. was supported by the Ministry of Economic Development of the Russian Federation (agreement with MIPT No. 139-15-2025-013, dated June 20, 2025, IGK 000000C313925P4B0002).
The funders had no role in study design, data collection and analysis, decision to publish, or preparation of the manuscript.

\bibliography{sample}

\section*{Legends}

\textbf{Figure 1.} Integration of expert knowledge into aleatoric uncertainty estimation provides total UE quality enhancement.

\textbf{Figure 2.} Overview of the proposed method. First, an ensemble of models is trained to predict ground truth labels from the dataset. Next, a different ensemble is created via fine-tuning on “soft” labels acquired from expert annotations. The two ensembles are then used to estimate epistemic and aleatoric uncertainty, respectively, given by equations \eqref{eq:epistemic} and \eqref{eq:aleatoric}. Finally, we evaluate the performance of algorithms with different uncertainty thresholds and construct rejection curves.

\textbf{Figure 3.} Rejection curves for proposed methods across four different machine learning tasks. Shaded areas represent the standard deviation between ensemble models. The higher the curve, the better.

\textbf{Figure 4.} \textbf{(a)} The pairwise consistency between experts estimated as the Cohen's kappa coefficient. Values on the diagonal are experts' accuracies. \textbf{(b)} Uncertainty estimation quality by the law of total variance where aleatoric uncertainty is estimated from different sources. \textit{Experts, best} stands for using the best $k$ experts based on accuracy, and \textit{Experts, average} is computed by averaging AAC for all possible sets of $k$ experts.

\textbf{Figure 5.} Relation of uncertainty estimation quality AAC on the multiple choice question answering task to the sizes of the base ensemble and the CAE. Lower values are better.

\textbf{Figure 6.} Accuracy that can be achieved with the Total Variance with CAE and a single model as the uncertainty estimation method for different rejection rates.

\textbf{Figure 7.} Accuracy-rejection curves compared between total variance methods using a base ensemble, an ensemble trained with a mixed loss, and a CAE on the BloodyWell dataset.

\textbf{Figure 8.} Accuracy-rejection curves compared between total variance methods using a base ensemble, an LLM as an oracle and experts on the PubMedQA dataset.

\textbf{Table 1.} Comparison of uncertainty estimation methods using rejection curves. AAC stands for the Area Above Curve and is normalized with respect to the optimal rejection-accuracy curve for a given model. The value in the bracket shows improvement over the baseline single-model method.

\textbf{Table 2.} Comparison of uncertainty estimation methods in error detection, where predicted uncertainty score is used as confidence in classification of model responses into errors and correct answers. AUROC stands for the Area Under the Receiver-Operator Curve. The best solution for each task is in bold and the second best solution is underlined.

\textbf{Table 3.} Dependency of AAC, $\times 10^{-3}$, on the weighting coefficient of epistemic uncertainty component for the Total Variance approach with CAE.

% \section*{Acknowledgments}

% This work was supported by the State Assignment of the Ministry of Science and Higher Education of the Russian Federation (Agreement No. FFNU-2025-0045).

\section*{Author contributions statement}

A.Y. and A.Z. provided theoretical derivations; A.Z. and E.E. conceived the experiments; A.K., E.Z., A.G. and S.K. conducted the experiments; A.K., E.Z., S.K., A.Z. and E.E. analyzed the results. All authors have reviewed the manuscript.

\section*{Additional information}

The authors declare no competing interests.
\end{document}

\flushbottom
\maketitle
% * <john.hammersley@gmail.com> 2015-02-09T12:07:31.197Z:
%
%  Click the title above to edit the author information and abstract
%
\thispagestyle{empty}

% \noindent Please note: Abbreviations should be introduced at the first mention in the main text – no abbreviations lists. Suggested structure of main text (not enforced) is provided below.

% \input{SP'2025 - Uncertainty/intro}

% \input{SP'2025 - Uncertainty/litreview}

% \input{SP'2025 - Uncertainty/uncerainty}

% \section*{Introduction}

% The Introduction section, of referenced text\cite{Figueredo:2009dg} expands on the background of the work (some overlap with the Abstract is acceptable). The introduction should not include subheadings.

% \begin{figure}[ht]
% \centering
% \includegraphics[width=0.7\linewidth]{images/frontal.pdf}
% \caption{Legend (350 words max). Example legend text.}
% \label{fig:frontal}
% \end{figure}

% \input{SP'2025 - Uncertainty/results}

% \subsection*{Subsection}

% Example text under a subsection. Bulleted lists may be used where appropriate, e.g.

% \begin{itemize}
% \item First item
% \item Second item
% \end{itemize}

% \subsubsection*{Third-level section}
 
% Topical subheadings are allowed.

% \input{SP'2025 - Uncertainty/discussion}

% \section*{Discussion}

% The Discussion should be succinct and must not contain subheadings.

\section*{Supplementary Methods}\label{sec:methods}

This section describes datasets used for evaluation, implementation details of the algorithms as well as details on baselines used in the Experiments section.
Additional justification for the mathematical model used in the paper is also provided below.

\subsection*{Datasets}

For this research, the datasets were required to contain annotations provided by multiple experts in order to generate the ``soft'' labels for confidence-aware training.
The final selection of benchmarks was curated to span various machine learning tasks and multiple domains.
% \todo[inline]{Better introduction to this section is needed. One solution would be to note that we specifically require datasets that contain expert annotations.}
% This section briefly describes datasets used in this work.

% \todo[inline]{No need in such thorough description of datasets.}

\subsubsection*{BloodyWell}

The BloodyWell dataset~\cite{korchagin2024image} comprises 3139 images depicting agglutination reactions, a process of erythrocytes clustering that is often used for blood typing. 
Each image is annotated with a ground truth binary label (positive or negative reaction result), independent assessments from six medical experts, and metadata including reagent types.
Instead of simple ``yes'' or ``no'' responses, experts were asked to provide a rating from 0 (high certainty that no agglutination is present) to 4 (high certainty that agglutination is present). 
The ground truth is reliable since all blood samples used belong to regular donors with repeatedly confirmed blood types.

% Experts provided independent evaluations based solely on the images and reagent types, simulating the information available to a neural network.
% This limitation aligns the experts' uncertainty more closely with the challenges faced by the neural network models.
% Nevertheless, the experts' classification performance is lower than in practice since this setup provides less contextual data (e.g., temperature, incubation time) than physical laboratory setting.

% For each reaction, expert assigned a score reflecting their certainty: ``Definitely positive'' (1.00), ``Most likely positive'' (0.75), ``Hard to classify'' (0.50), ``Most likely negative'' (0.25), ``Definitely negative'' (0.00).
% The complexity of the agglutination reactions varies due to reagent properties, individual blood sample characteristics, environmental conditions, and the dynamics of the physical process within the well.
% Votes from all six experts are in complete agreement for over 70\% of the reactions, verifying reliability of their annotations. 

\subsubsection*{RIGA}
The RIGA~\cite{almazroa2018retinal} dataset contains images of retinal fundus annotated by six experienced ophthalmologists.
The data is split into three subsets named “MESSIDOR”, “Bin Rushed” and “Magrabi” containing 460, 195 and 95 images respectively for a total of 750 images.
Each image is of size 256 × 256 pixels and is supplied with six masks of an optic disc and an optic cup, one for each expert.
Ophthalmologists often use the vertical and horizontal cup-to-disk ratios as well as the disk and cup area ratios for diagnosing glaucoma.

\subsubsection*{LIDC}
The LIDC-IDRI~\cite{armato2011lung} contains 1018 lung CT scans annotated by 4 radiologists for nodule segmentation with binary maps.
% Consistency of their responses in each pixel is considered to be average confidence.
In our experiments, volumetric scans were divided into 2D slices, which were then cropped to images of size 128 × 128 centered around the nodule annotation.
In total, 15096 images were acquired.

\subsubsection*{PubMedQA}
PubMedQA dataset~\cite{jin2019pubmedqa} consists of medical questions derived from the titles of medical research articles from the PubMed database.
The dataset was constructed by exclusively including articles with abstracts containing an explicitly labeled ``Conclusion'' section.
Therefore, each question is annotated with a long answer (the "Conclusion" section of the abstract) and context (the remainder of the abstract).
The dataset has three subsets: manually labeled (1,000 samples), automatically labeled (211,300 samples), and unlabeled (61,200 samples).
The manual labeling was performed by two medical experts, both of which had access to the ``context'' section and only one expert having access to the ``long answer'' section of the sample.
They independently provided a short answer for each question (``yes'', ``no'', or ``maybe'') on the basis of the context.
The experts then had a discussion and reached the consensus ground truth label.

\subsection*{Implementation Details}

% \todo[inline]{No need in duplication of experimental setups for already published results. }

This section provides technical details on the experimental setup, including data splits, model architectures, data augmentation as well as chosen parameters for training and inference.

For all considered tasks, the sizes of both the base ensemble and the CAE are 10 models.
Model diversity was ensured by random weight initializations and a cross-validation training procedure, in which the overall training data was uniquely split into training and validation subsets for each model.
Since ground truth segmentation maps were not available for both RIGA and LIDC-IDRI tasks, a random expert was chosen as a ground truth.
The soft labels were generated by simply averaging all available expert predictions.
More advanced methods like Spatially Varying Label Smoothing~\cite{islam2021spatially} for segmentation tasks were considered but we observed no substantial improvement in uncertainty estimation quality.

The models achieved the accuracy of $96.71\% \pm 0.58\%$ in image classification and $71.18\%\pm0.85\%$ in multiple-choice question answering, as well as Dice scores of $0.6348\pm0.0229$ for binary segmentation and $0.8638\pm0.0081$ for multiclass segmentation.

\subsubsection*{BloodyWell}
The dataset was split in training, validation, and test with a ratio of 1:1:8.

We selected MobileNet-V3-small architecture~\cite{howard2019searching} as it is simple yet effective for image classification tasks. 
To account for the domain specificity, the reagent type, a factor known to significantly influence the reaction process and available to laboratory assistants, was provided as an additional input during training.
The reagent type was encoded as a one-hot vector and concatenated with the features at the input of the final fully connected layer.

Optimization was performed using the Adam optimizer~\cite{kingma2014method} with parameters $\beta_1 = 0.9$ and $\beta_2 = 0.999$ and a binary cross-entropy loss function. 
The base ensemble was trained for 400 epochs with an initial learning rate of $4 \cdot 10^{-4}$, which was reduced by a factor of 0.8 every 25 epochs.
A weight decay coefficient of $5 \cdot 10^{-5}$ was applied for regularization.
To improve model generalization and robustness, a plethora of data augmentations were applied, including random rotations, crops, flips, projective transformation, contrast change, sharpness change and addition of Gaussian noise.

The CAE was obtained by fine-tuning the base ensemble for another 300 epochs.
The mixed-loss ensemble was trained similarly to the base ensemble, but a mixed loss function was employed with a weighting coefficient of $\alpha = 0.9$. The selection between ``hard'' and ``soft'' labels for each data point was performed with equal probability.

The \textit{MCMC}~\cite{welling2011bayesian} training prodecure consisted of fine-tuning the base ensemble with the Stochastic Gradient Langevin Dynamics (SGLD) optimizer for 150 epochs. 
Model checkpoints were saved every 15 epochs. 
Uncertainty was quantified as the predictions variance across the checkpoints generated from the corresponding initial model.
For the \textit{MC Dropout}~\cite{gal2016dropout} approach, the dropout probability at inference was set to 20\%, matching the training phase. 
For each input, 25 stochastic forward passes were performed, and the uncertainty was calculated as the variance of the resulting predictions.

In the \textit{HUQ}~\cite{vazhentsev2023hybrid} method, epistemic and aleatoric uncertainty components were quantified according to the DDU~\cite{mukhoti2023deep} approach. 
Epistemic uncertainty was computed as the density in the model feature space with features obtained as Gaussian Mixture Models (GMM) on 10-component PCA-reduced network embeddings extracted after the final convolutional layer.
Aleatoric uncertainty was calculated as the cross-entropy of the network output. 
The ranking function was trained on 10\% of the validation set with a trade-off hyperparameter $\alpha$ set to $0.5$.

%%%%%%%%
Finally, for the \textit{Test-time Augmentations}~\cite{wang2018test}, we used the same augmentation techniques and corresponding parameters as during training. 
For each image, 25 modified versions were generated.
Uncertainty was quantified as the variance of model predictions on these augmented inputs.

% \textbf{Conformal Prediction Sets}~\cite{messoudi2020deep}. 
% The validation set served as the calibration one. 
% The distance metric for identifying neighbors was the L2-norm applied to network embeddings derived from the last convolutional layer and reduced using a 10-component PCA.
% Uncertainty for a sample was calculated as the sum of absolute differences between the model's prediction and the true labels for its 60 nearest neighbors within the calibration set.

\subsubsection*{LID-IDRI and RIGA}

The training procedure for both segmentation tasks was identical.
The model of choice was a standard UNet~\cite{ronneberger2015u} with Batch Normalization layers~\cite{ioffe2015batch} as a strong baseline validated in a large sample of similar problems, with only differences between tasks in the first and the final layer to account for the data format.
Models were trained with the Adam~\cite{kingma2014method} optimizer with $\beta_1 = 0.9$ and $\beta_2 = 0.999$, an initial learning rate of $10^{-4}$ decayed by a factor of 0.9 every 10 epochs, and a weight decay of $10^{-4}$.
The models in the base ensemble were trained from random parameter initialization for 100 epochs, and the CAE models were fine-tuned from the base ensembles for another 100 epochs.
The loss function of choice was cross-entropy.

For the LIDC-IDRI task, the data was divided in proportion of 6:2:2 into the train, validation and test splits.
For the RIGA task, the Bin Rushed subset was used as a test set while Magrabi and MESSIDOR subsets were combined and randomly separated in proportion of 3:1 into the training and the validation parts.
Random flips and rotations were applied as augmentations to enrich the data.

Training regimes for MCMC models follow theones outlined for the BloodyWell dataset.
ABNN~\cite{franchi2024make} models were fine-tuned for another 100 epochs for the ease of comparison, and each model was sampled 10 times, with variance between predictions serving as the uncertainty score.

\subsubsection*{PubMedQA}

We trained an ensemble of 10 ModernBERT~\cite{modernbert} models on the final decision labels in 3 stages:

\begin{itemize}
    \item For each model, randomly sample 70\% entries from the artificial subset. Train in ``no reasoning required'' setting: pass the context with question without providing the detailed answer. Train for 1 epoch and use $10^{-6}$ learning rate.
    \item For each model, use the same corresponding sample from the artificial subset. Train in ``reasoning required'' setting: pass the context with question and the detailed answer. Train for 1 epoch and use $10^{-6}$ learning rate.
    \item Split the training part of the expert-labelled subset into 10 folds. Train each model on 9 different folds. Fine-tune in ``reasoning required'' setting: pass the context with question and the detailed answer. Train for 5 epoch and use $10^{-6}$ learning rate.
\end{itemize}

We also trained an ensemble of 10 ModernBERT models with the same hyperparameters on the final decision labels for the first 2 stages and the expert average prediction for the final stage.

Implementations of MC Dropout and HUQ follow those outlined in the BloodyWell section.
Test-time augmentation was performed by first generating 10 synonymous yet distinct versions of the original question via prompting the Mistral Medium model of version 25.05.
The trained ModernBERT ensemble was then used to generate predictions, and the variance of predictions was taken as the final uncertainty estimate.

% Detailed list of hyperparameters can be found in the repo.

\subsection*{Generating confidence scores from an LLM ``oracle''}
% To generate confidence scores from an LLM, the Mistral Medium model of version 25.05 was prompted with the following query:\\
% \texttt{
% Here is the question:\\
% \{question\}\\
% Please provide your best answer guess (choosing from "yes", "no", "maybe") and your confidence in it from 0\% (absolutely uncertain) to 100\% (absolutely certain)\\ 
% in the following JSON format:\\
% {{\\
% "answer": "Your answer here",\\
% "confidence\_score": number\\
% }}\\
% }

To generate confidence scores from an LLM, the Mistral Medium model of version 25.05 was prompted with the following query:
\begin{verbatim}
Here is the question:
{question}
Please provide your best answer guess (choosing from "yes", "no", "maybe")
and your confidence in it from 0% (absolutely uncertain)
to 100% (absolutely certain) in the following JSON format:
{{
"answer": "Your answer here",
"confidence_score": number
}}
\end{verbatim}

Ten answers for each sample were acquired, and confidence scores averaged to achieve the final per-sample score.
Although more complex prompts were tested to generate responses that better estimate aleatoric uncertainty, we were unable to improve the quality of this approach.

\subsection*{Rationale for multi-class uncertainty aggregation}
In this section, we share intuition behind viewing aggregate uncertainty in multi-class segmentation as simply a sum of uncertainties from different classes.
To do so, we model the target $y$ as a random $N$-dimensional vector; for each $n \in \{1, \ldots, N\}$, the $n^{\text{th}}$ component of $y$ is equal to $1$ if and only if sample $x$ belongs to class $n$, and is $0$ otherwise.
For a sample $x$, vector $y$ is modeled in terms of a categorical distribution with $N$ categories: $y\mid x\sim \mathrm{Cat}(N, \mathbf{p}(x))$.
The model $\mathbf p_{\theta}(x)$ is then trained to predict the vector-column $\mathbf p(x)$ such that, for any $n \in \{1, \ldots, N\}$, $[\mathbf p_{\theta}(x)]_n = \mathbb{P}(y_n = 1\mid x, \theta) \approx \mathbb{P}(y_n = 1\mid x) = [\mathbf p(x)]_n$; here, $\mathbf v_m$ refers to the $m^{\text{th}}$ component of vector $\mathbf v$.

An expected value and a covariance matrix of a categorical distribution are expressed in terms of $\mathbf p(x)$: $\mathbb{E} (y \mid x) = \mathbf p(x)$, $\mathrm{Var}(y\mid x) = \mathrm{diag}(\mathbf p(x))-\mathbf p(x)\mathbf p(x)^\top.$
We can then derive the Monte-Carlo estimates of uncertainty components similar to binary classification:

\begin{align}
    \mathrm{Var}_{\theta\sim \mathbb{P}(\theta)} \left[\mathbb{E}(y \mid x, \theta)\right] &\approx \frac{1}{K}\sum_{i=1}^{K}\left(\mathbf p_{\theta_i}(x) - \mathbf {\overline{p}}_{\theta}(x)\right)\left(\mathbf p_{\theta_i}(x) - \mathbf {\overline{p}}_{\theta}(x)\right)^\top, \label{eq:epistemic-mul} \\ 
    \mathbb{E}_{\theta\sim \mathbb{P}(\theta)}\left[\mathrm{Var}(y \mid x, \theta)\right] &\approx \frac{1}{K}\sum_{i=1}^{K} \left( \mathrm{diag}(\mathbf p_{\theta_i}(x))-\mathbf p_{\theta_i}(x)\mathbf p_{\theta_i}(x)^\top\right). \label{eq:aleatoric-mul}
\end{align}

Unlike binary classification, total variance is now represented as a matrix instead of a scalar, which complicates further analysis.
A covariance matrix $V$ is sometimes replaced with $\mathrm{det} (V)$ or $\mathrm{tr} (V)$ to represent aggregate covariance with a scalar.
Finally, we note that
\begin{equation}
    \mathrm{tr}\left[\mathrm{Var}(y\mid x)\right] =
    \mathrm{tr}\left[\mathrm{Var}_{\theta \sim \mathbb{P}(\theta)}[\mathbb{E}(y\mid x, \theta)]\right] + \mathrm{tr}\left[\mathbb{E}_{\theta\sim\mathbb{P}(\theta)}[\mathrm{Var}(y\mid x, \theta)]\right],
\end{equation}
and
\begin{align}
    \mathrm{EU}(y) 
    &= \mathrm{tr}\left[\mathrm{Var}_{\theta \sim \mathbb{P}(\theta)}[\mathbb{E}(y\mid x, \theta)]\right]
    \approx \mathrm{tr}\left[\frac{1}{K}\sum_{i=1}^{K}\left(\mathbf p_{\theta_i}(x) - \mathbf {\overline{p}}_{\theta}(x)\right)\left(\mathbf p_{\theta_i}(x) - \mathbf {\overline{p}}_{\theta}(x)\right)^\top\right] \notag \\ 
    &= \sum_{j = 1}^N \left[ \frac1K \sum_{i = 1}^K ([\mathbf p_{\theta_i}(x)]_j - [\mathbf{\overline p}_{\theta}(x)]_j)^2 \right] \notag  
    = \sum_{j = 1}^N \mathrm{EU}(y_j),
\end{align}
\begin{align}
    \mathrm{AU}(y) &= \mathrm{tr}\left[\mathbb{E}_{\theta\sim\mathbb{P}(\theta)}[\mathrm{Var}(y\mid x, \theta)]\right]
    \approx \mathrm{tr}\left[\frac{1}{K}\sum_{i=1}^{K} \left( \mathrm{diag}(\mathbf p_{\theta_i}(x))-\mathbf p_{\theta_i}(x)\mathbf p_{\theta_i}(x)^\top\right)\right] = \notag \\
    &= \sum_{j = 1}^N \left[ \frac1K \sum_{i = 1}^K ([\mathbf p_{\theta_i}(x)]_j - [\mathbf{p}_{\theta_i}(x)]^2_j) \right] \notag
    = \sum_{j = 1}^N \mathrm{AU}(y_j),
\end{align}
which leads to
\begin{equation}
    \mathrm{TU}(y) = \mathrm{EU}(y) + \mathrm{AU}(y) \approx \sum_{j=1}^N\mathrm{EU}(y_j)+\sum_{j = 1}^N\mathrm{AU}(y_j) = \sum_{j = 1}^N \mathrm{TU}(y_j).
\end{equation}
Therefore, the proposed uncertainty aggregation approach is not only convenient but also naturally emerges from a probabilistic modeling of the problem at hand.

\bibliography{sample}

% For data citations of datasets uploaded to e.g. \emph{figshare}, please use the \verb|howpublished| option in the bib entry to specify the platform and the link, as in the \verb|Hao:gidmaps:2014| example in the sample bibliography file.

% \section*{Acknowledgments (not compulsory)}

% Acknowledgments should be brief, and should not include thanks to anonymous referees and editors, or effusive comments. Grant or contribution numbers may be acknowledged.

% \section*{Author contributions statement}

% Must include all authors, identified by initials, for example:
% A.A. conceived the experiment(s),  A.A. and B.A. conducted the experiment(s), C.A. and D.A. analysed the results.  All authors reviewed the manuscript. 

% \section*{Additional information}

% To include, in this order: \textbf{Accession codes} (where applicable); \textbf{Competing interests} (mandatory statement). 

% The corresponding author is responsible for submitting a \href{http://www.nature.com/srep/policies/index.html#competing}{competing interests statement} on behalf of all authors of the paper. This statement must be included in the submitted article file.

% \begin{figure}[ht]
% \centering
% \includegraphics[width=\linewidth]{stream}
% \caption{Legend (350 words max). Example legend text.}
% \label{fig:stream}
% \end{figure}

% \begin{table}[ht]
% \centering
% \begin{tabular}{|l|l|l|}
% \hline
% Condition & n & p \\
% \hline
% A & 5 & 0.1 \\
% \hline
% B & 10 & 0.01 \\
% \hline
% \end{tabular}
% \caption{\label{tab:example}Legend (350 words max). Example legend text.}
% \end{table}

% Figures and tables can be referenced in LaTeX using the ref command, e.g. Figure \ref{fig:stream} and Table \ref{tab:example}.